%% file: main.tex
\documentclass[twoside]{article}

\usepackage[accepted]{aistats2024}
\input{math_commands.tex}

\input{math_alan}

\usepackage[utf8]{inputenc}
\usepackage[T1]{fontenc}
\usepackage{paralist}            % compact enumerate
\usepackage{xcolor}              % color
\usepackage{mathtools}           % comment under arrow
\usepackage{comment}             % comment out
\usepackage{soul}
\usepackage{makecell}            % line break within table
\usepackage{minitoc}
\usepackage{pifont}
\usepackage{booktabs}
\usepackage{hyperref}
\newcommand{\xmark}{\textcolor{gray!50}{\ding{55}}}
\usepackage[noabbrev,nameinlink]{cleveref}
\usepackage{braket}
\usepackage{url}            
\usepackage{graphicx}    
\usepackage{booktabs}  
\usepackage{algorithm, algpseudocode}%
\floatname{algorithm}{Algorithm}

\usepackage{amsthm}

\algtext*{EndFunction}
\usepackage{amsmath}
\usepackage{bm}
\usepackage{amsfonts} 
\usepackage{dsfont}
\usepackage{wrapfig}
\usepackage{graphicx}
\usepackage{natbib}
\usepackage{amsthm}
\usepackage{caption}
\usepackage{subcaption}
\usepackage{caption}
\usepackage{amssymb}
\usepackage{wrapfig}
\usepackage[normalem]{ulem}
\usepackage{nicefrac}       % compact symbols for 1/2, etc.
\usepackage{microtype}      % microtypography
\usepackage{lipsum}
\usepackage{graphicx}
\usepackage{caption}
\usepackage{xcolor}
\usepackage{upgreek}
\usepackage{thmtools}
\usepackage{thm-restate}
\usepackage{amsmath}
\usepackage[inline]{enumitem}

\newtheorem{defn}{Definition}

\newtheorem{theorem}[defn]{Theorem}
\newtheorem{lemma}[defn]{Lemma}
\newtheorem{proposition}[defn]{Proposition}

\DeclareFontFamily{U}{solomos}{}
\DeclareErrorFont{U}{solomos}{m}{n}{10}
\DeclareFontShape{U}{solomos}{m}{n}{
  <-> s*[1.1]  gsolomos8r
}{}

\begin{document}

\newcommand{\Alan}[1]{\textcolor{blue}{\textbf{Alan}: #1}}
\newcommand{\Masaki}[1]{\textcolor{red}{\textbf{Masaki}: #1}}
\newcommand{\Seb}[1]{\textcolor{gray}{\textbf{Sebastian}: #1}}

\twocolumn[

\aistatstitle{Looping in the Human: \\ Collaborative and Explainable Bayesian Optimization}
\aistatsauthor{
    \begin{tabular}{@{}c@{}}
    Masaki Adachi$^{1,2,3}$ \qquad 
    Brady Planden$^{2}$ \qquad
    David A. Howey$^{2,4}$ \qquad 
    Michael A. Osborne$^{1,2}$ \\
    Sebastian Orbell$^{2}$ \qquad
    Natalia Ares$^{2}$ \qquad
    Krikamol Muandet$^{5}$ \qquad 
    Siu Lun Chau$^{5}$
    \end{tabular}
}
\aistatsaddress{
$^{1}$Machine Learning Research Group, University of Oxford, OX2 6ED, United Kingdom\\
$^{2}$Department of Engineering Science, University of Oxford, OX1 3PJ, United Kingdom\\
$^{3}$Toyota Motor Corporation, Shizuoka 410-1107, Japan \\
$^{4}$The Faraday Institution, Harwell Campus, Didcot OX11 0RA, United Kingdom\\
$^{5}$CISPA Helmholtz Center for Information Security, 66123 Saarbr\"ucken, Germany \\
}
]

\begin{abstract}
Like many optimizers, Bayesian optimization often falls short of gaining user trust due to opacity. While attempts have been made to develop human-centric optimizers, they typically assume user knowledge is well-specified and error-free, employing users mainly as supervisors of the optimization process. We relax these assumptions and propose a more balanced human-AI partnership with our Collaborative and Explainable Bayesian Optimization (CoExBO) framework. Instead of explicitly requiring a user to provide a knowledge model, CoExBO employs preference learning to seamlessly integrate human insights into the optimization, resulting in algorithmic suggestions that resonate with user preference. CoExBO explains its candidate selection every iteration to foster trust, empowering users with a clearer grasp of the optimization. Furthermore, CoExBO offers a no-harm guarantee, allowing users to make mistakes; even with extreme adversarial interventions, the algorithm converges asymptotically to a vanilla Bayesian optimization. We validate CoExBO's efficacy through human-AI teaming experiments in lithium-ion battery design, highlighting substantial improvements over conventional methods. Code is available \url{https://github.com/ma921/CoExBO}.
\end{abstract}
\vspace{-1em}

\doparttoc % Tell to minitoc to generate a toc for the parts
\faketableofcontents % Run a fake tableofcontents command for the partocs
%new skeleton
\vspace{-1em}
\section{Introduction}
\begin{figure}[hbt!]
    \centering
    \includegraphics[width=0.5\textwidth]{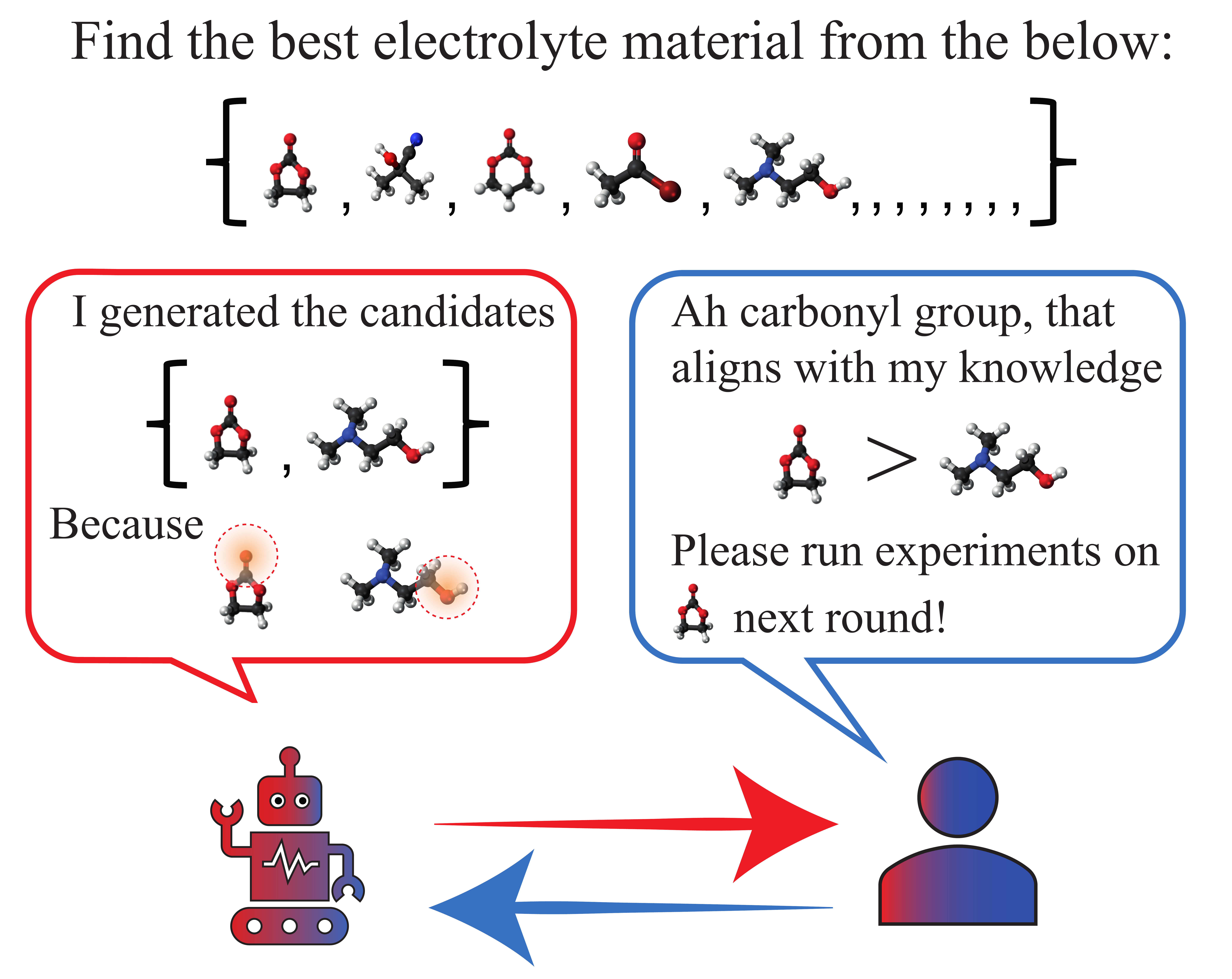}
    \caption{In Collaborative and Explainable Bayesian Optimization (CoExBO), a human expert collaborates with BO to refine electrolyte materials. While experts excel in discerning material differences rather than identifying the best one, pairwise comparisons and explanations boost their feedback accuracy and trust. This guides the BO to produce better candidates, ensuring quicker convergence.}
    \label{fig:concept}
    \vspace{-1em}
\end{figure}
\begin{figure}[hbt!]
  \centering
  \includegraphics[width=1\hsize]{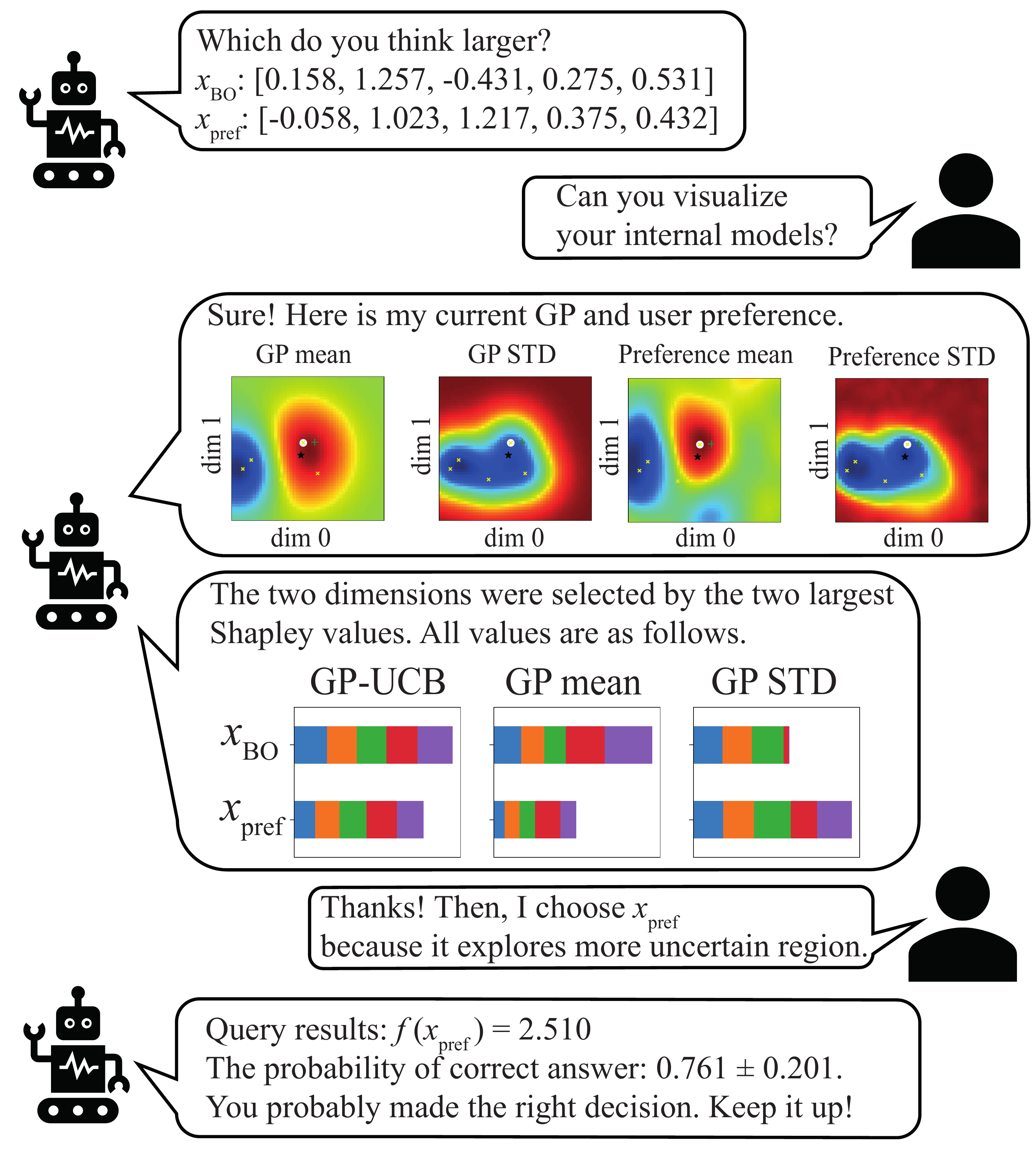}
  \caption{Explanation flow: \textbf{Spatial relation:} BO visualizes the surrogate model's predictive distribution and estimated human preference models for the two primary dimensions determined by Shapley values. \textbf{Feature importance:} Users' values are provided for both candidates' predictive mean, standard deviation, and acquisition function. \textbf{Selection accuracy feedback:} After observing the function value, a post-hoc evaluation of the correct selection probability is given.}
  \label{fig:explanation}
  \vspace{-2em}
\end{figure}
Bayesian optimization (BO) is a popular blackbox optimizer for expensive-to-evaluate tasks. While it is widely applied in diverse domains \citep{feurer2015efficient, wu2020practical, adachi2021high}, it has yet to fully gain human users' trust. Surveys from NeurIPS2019/ICLR2020~\citep{bouthillier2020survey} found that most AI researchers prefer manually tuning hyperparameters. This is surprising given that \citet{bergstra2012random} has shown manual search performs worse than simple random search and lacks global convergence guarantees \citep{gupta2023bo}.

To make BO trustworthy, recent research has moved towards human-AI collaborative paradigms \citep{kanarik2023human}. These methods often make contrasting assumptions about human exploration abilities. Suppose humans are superior to BO \citep{colella2020human, av2022human}, their intervention can enhance convergence---but this potent assumption also implies manual search would be superior to BO, undermining the core justification for using BO. 
Conversely, if humans are viewed as imperfect agents, existing works such as \citet{gupta2023bo, khoshvishkaie2023cooperative} treat humans as a central optimizer, and BO supports human manual search via exploratory adjustment, that can assure the global convergence even with erroneous human selection. When one side is better, the inferior side's selection is wasteful, leading to a worse convergence rate than vanilla BO \citep{khoshvishkaie2023cooperative}.

To build a balanced human-BO partnership, we believe an ideal method should satisfy the following criteria:
(a) \textbf{Explainability:} The method should enhance user understanding of the optimization process, promoting transparency. While \citet{li2020explainability} introduced explainability by limiting the search space, it may overly restrict it, lacking a global convergence guarantee.
%\Alan{what's the limitation of their work? Otherwise if it is concurrent work, we put it in conclusion. Just putting a sentence breaks the flow.}
(b) \textbf{User-centric knowledge integration:} An ideal approach should seamlessly incorporate human insights from user interactions without requiring users to define an exact knowledge model. For instance, materials often exist in a high-dimensional feature space, whereas BO typically uses a reduced low-dimensional feature set due to limited chemistry data \citep{jordan2019artificial}. Chemists have discernment to assess materials using information inaccesible to the model but struggle to articulate it quantitatively \citep{cisse2023hypbo}. Hence, users face challenges with existing methods; e.g., \citet{hvarfner2022pi} mandates a prior function capturing the users' optimal location belief, while \citet{av2022human} requires users to select the next query point quantitatively. In short, knowledge elicitation in high-dimensional domains is notoriously intricate \citep{rousseau2001schema, garthwaite2005statistical, mikkola2021prior}.
(c) \textbf{Robustness:} The method should be robust against human errors, offering a no-harm guarantee to ensure that even adversarial interventions do not adversely impact the vanilla BO convergence rate. 
To the best of our knowledge, only \citet{hvarfner2022pi} can theoretically assure the no-harm guarantee.
Crucially, none encompass all three comprehensively.

This paper introduces the \emph{Collaborative and Explainable Bayesian Optimization }(CoExBO) framework to tackle the above challenges. For criterion (a), CoExBO employs Shapley values~\citep{shapley1953value}, a cornerstone of explainable AI, to ensure users can effectively understand and interpret the candidate acquisition mechanism. Addressing criterion (b), CoExBO deviates from conventional methods that require users to input an exact knowledge model. Instead, it presents users with candidate pairs, empowering them to select the perceived optimal one. This approach allows CoExBO to implicitly assimilate human insights via preference learning~\citep{bradley1952rank}. This is grounded that humans excel at relative comparisons rather than quantifying an absolute preference for a singular choice~\citep{kahneman1979interpretation}. For criterion (c), inspired by \citet{hvarfner2022pi}, our candidate generation strategy prioritizes expert knowledge in the early optimization stages. As more experimental data accumulates and refines the surrogate model, the influence of human input gradually diminishes. Theorem \ref{thm:regret} proves this offers a no-harm guarantee.

Our contributions are summarized as:
\begin{compactenum}
    \item We introduce CoExBO, a novel framework promoting a balanced human-BO partnership. CoExBO is characterized by its transparency, capacity to assimilate human insights seamlessly, and resilience against human errors.
    \item We establish the efficacy of CoExBO via real-world optimization tasks on lithium-ion battery design problems, and the expert-BO team can gain significant speedup over eight popular baselines.
\end{compactenum}

\section{Bayesian optimization and existing human-in-the-loop extensions}

\textbf{Bayesian optimization.}
We aim to optimize the function $f$ 
%given no access to its functional form nor its gradient---
when $f$ can only be queried pointwise. %We wish to find
\begin{align}
    x^*_\text{true} = \mathop\mathrm{argmax}_{x \in \mathcal{X}} f(x), \label{eq:obj_bo}
\end{align}
where $\mathcal{X} \subseteq \RR^d$ is the $d$ dimensional continuous input domain, and $x^*_\text{true}$ is the global optimum. We assume that $f(x)$ is costly to query and can only observe a noisy estimate $y = f(x) + \epsilon$ with i.i.d. zero-mean Gaussian noise $\epsilon$. The goal is to find the optimal $f(x)$ under a given number of queries.

BO \citep{mockus1998application, garnett2023bayesian} is a model-based black box optimizer that employs a Gaussian process (GP) \citep{williams2006gaussian} as a surrogate model. At optimization step $t$, the GP approximates the true function $f$ using the current observations ${\textbf{D}_\text{obj}}_t = ({\bfX_\text{obj}}_t, {\bfy_\text{obj}}_t)$ as $f_t \sim \cG\cP(\mu_t, \kappa_t)$, with
\begin{align*}%\label{eq:GP}
    \mu_t(x) &= k(x, {\bfX_\text{obj}}_t){\bfK^{\prime \, -1}_{\bfX\bfX}}_t {\bfy_\text{obj}}_t,\\
    \kappa_t(x, x^\prime) &= k(x, x^\prime) - k(x, {\bfX_\text{obj}}_t){\bfK^{\prime \, -1}_{\bfX\bfX}}_t k({\bfX_\text{obj}}_t, x),
\end{align*} 
where $\mu_t$ and $\kappa_t$ are the GP's posterior predictive mean and covariance functions at round $t$. $k$ is the kernel, $\lambda > 0$ is the Gaussian likelihood variance, ${\bfK_{\bfX\bfX}}_t := k({\bfX_\text{obj}}_t, 
{\bfX_\text{obj}}_t)$, ${\bfK^{\prime \, -1}_{\bfX\bfX}}_t := ({\bfK_{\bfX\bfX}}_t + \lambda \textbf{I})^{-1}$, and $\textbf{I} \in \mathbb{R}^{d \times d}$ is the identity matrix. Using GP predictive uncertainty, BO solves the blackbox optimization problem as active learning and selects the next query point by maximising an acquisition function (AF). One popular class of AF is the upper confidence bound (UCB) \citep{srinivas2009gaussian}: ${\alpha_{f_t}}(x) := \mu_t(x) + \beta^{1/2}_t \sigma_t(x),$
% \begin{align*}
%     {\alpha_{f_t}}(x) := \mu_t(x) + \beta^{1/2}_t \sigma_t(x),%\label{eq:ucb}
% \end{align*}
where $\sigma_t(x) := \sqrt{\kappa_t(x,x)}$ is the standard deviation of the GP predictive posterior, $\beta_t$ is the user-specified parameter indicating the trade-off between exploitation~(using current knowledge of optimum from $\mu_t$) and exploration~(considering the uncertainty from $\sigma_t$). %In the following, when the context is clear, we drop the subscript $t$ to avoid notation overflow.

\textbf{Human-in-the-loop extensions.} %To the best of our knowledge, 
There are four prevailing approaches to integrate human knowledge into BO: (1) By treating human knowledge as a prior over the input space \citep{souza2021bayesian, ramachandran2020incorporating, hvarfner2022pi, cisse2023hypbo}. (2) By adopting a hyperprior over the function space \citep{hutter2011sequential, snoek2014input, wang2021pre}. (3) By considering it as a multi-fidelity information source \citep{song2019general, huang2022bayesian}. (4) Implementing human knowledge as hard constraints \citep{gelbart2014bayesian, hernandez2015predictive, adachi2024adaptive}. Among these categories, only the work of \citet{hvarfner2022pi}, belonging to the first category, provides a no-harm guarantee against potential human errors. Notably, all these methods operate under the assumption that human knowledge can be well specified to the algorithm. A more detailed related work section is delineated in Supplementary \ref{sup:related}.

\textbf{$\pi$BO.} CoExBO is inspired by the $\pi$BO algorithm \citep{hvarfner2022pi}, which characterizes human knowledge as a prior distribution representing their belief in the global optimum location, i.e., $\pi(x) :=  \mathbb{P} \big(f(x) = \max_{x' \in \mathcal{X}} f(x')
    \big).$
% \begin{align*}
%     \pi(x) &:=  \mathbb{P} \big(f(x) = \max_{x' \in \mathcal{X}} f(x')
%     \big). %\label{eq:pi_def}
% \end{align*}
This prior can then be incorporated into an AF $\alpha_t$ to act as a soft constraint for a warmer start on the optimization. Specifically, at round $t$, we search for $x_\text{next} = \mathop\mathrm{argmax}_{x \in \mathcal{X}} \ \alpha_t (x) \pi(x)^{\gamma / t}$
where $\gamma > 1$ controls the decay rate of this constraint. This decay of human contribution is justified as follows: at the start of the BO, expert knowledge can help substantially, whereas, at later stages, the BO will likely have enough data to reach the optimum confidently. Furthermore, this decaying property is the key reason behind the no-harm guarantee in Corollary 1 in \citet{hvarfner2022pi}.

While $\pi$BO's formulation is straightforward, requiring the user to specify a prior over high-dimensional input space could be very challenging in practice~\citep{garthwaite2005statistical}. The following section demonstrates how CoExBO can relax this assumption by interacting with users through preference elicitation.

% \vspace{-1em}
\section{Collaborative and Explainable BO}
In this section, we present our \underline{Co}llaborative and \underline{Ex}plainable \underline{B}ayesian \underline{O}ptimization (CoExBO) algorithm. While its objective aligns with the conventional BO objective (Eq.~\ref{eq:obj_bo}), CoExBO differentiates itself by explaining the acquisition process and incorporating human knowledge through preference learning. Specifically, the query procedure consists of the following steps: At round $t > 1$ with surrogate GP $f_t$ and preference model $\hat{\pi}_{t}$, we have
\begin{align*}
    \Gamma(f_t, \hat\pi_t) &\rightarrow (x_1, x_2), \tag{\small{Acquire candidates}}\\
    \operatorname{E}(f_t, x_1, x_2) &\rightarrow (\boldsymbol{\phi}_1, \boldsymbol{\phi}_2)  \tag{\small{Explain acquisition}}\\
    \operatorname{H}\left(\{(x_i, \boldsymbol{\phi}_i)\}_{i=1}^2\right) &\rightarrow \tilde{x} \in (x_1, x_2), \tag{\small{Elicit preference}} \\
    \mathbf{\Pi}\left(\hat{\pi}_t, \tilde{x} ,x_1, x_2 \right) &\rightarrow \hat{\pi}_{t+1}, \tag{\small{Update} $\hat{\pi}_t$}
\end{align*}
and finally we run the experiment with $\tilde{x}$ to obtain $y_{\text{next}} = f(\tilde{x})$. We denote $\Gamma$ as the candidate generation function (see $\S \ref{subsec:candidate_generation}$) that takes in the surrogate and current preference models and generates a pair of candidates. $\operatorname{E}$ is an explanation function that explains the acquisition process (see \S\ref{subsec:explanation}) and returns explanation $\boldsymbol{\phi}_i$ for the $i^\text{th}$ candidate. $\operatorname{H}$ denotes the human users' choice when pairs of candidates and the subsequent explanations are given. Preference function $\hat{\pi}_t$ is then updated to $\hat{\pi}_{t+1}$ by taking into account the human preference through an update function $\mathbf{\Pi}$ (see \S \ref{subsec:preference_learning}), and at last we end the iteration by running an experiment on the chosen candidate $\tilde{\boldsymbol{x}}$. %In the following subsections, we delve deeply into each integral component of CoExBO. 

\subsection{Model human knowledge through preference learning}
\begin{figure}
  \centering
  \includegraphics[width=0.7\hsize]{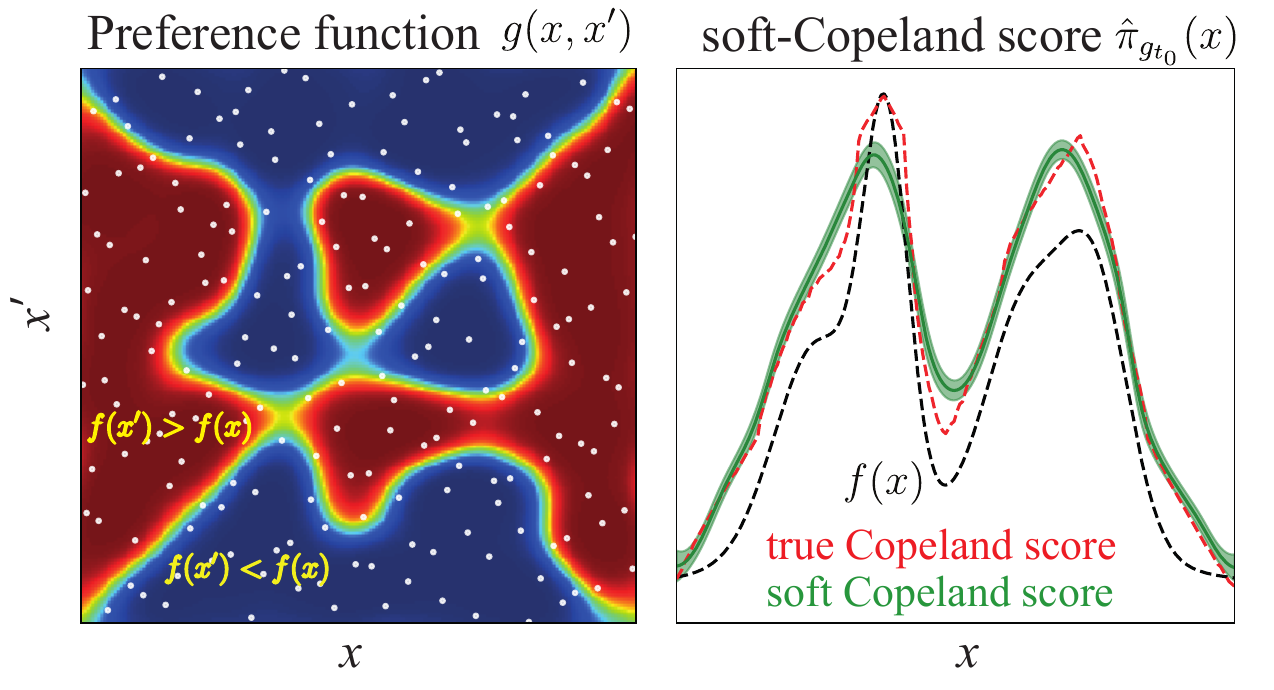}
  \caption{Preference learning concepts: we aim to model the ordinal relationship $f(x) < f(x^\prime)$ and its inverse with GP, utilizing the dataset $D_\text{pref}^{t_0}$, represented by white dots. Soft-Copeland score is used for the proxy of true function estimate.%We model a preference function $g(x,x^\prime)$ by a Gaussian process. By integrating this function, we compute the soft-Copeland score, denoted as $\hat{\pi}_{g_{t_0}}(x)$.
  }
  \label{fig:copeland}
  \vspace{-1em}
\end{figure}
\label{subsec:preference_learning}
While $\pi$BO requires the user to provide the preference function $\pi$ explicitly---which might be challenging to elicit in practice---we relax this assumption and estimate $\pi$ by $\hat{\pi}:\cX\to\RR_{+}$ using preference learning. At its core, preference learning aims to model the order relationships among a candidate set, $\cX$. For this paper, our emphasis is on binary preference learning, as detailed in \citep{bradley1952rank,chau2022learning}. However, this concept can be straightforwardly extended to more complex preference models such as choice functions~\citep{benavoli2023learning}.

To initiate the optimization process, we randomly select candidate pairs from $\cX$, then solicit the users' opinion on which one is more likely the optimal location. Formally, at $t=t_0$, we sample $J_{t_0}$ binary comparisons, denoted as $D_{\text{pref}}^{t_0} := \{x_1^{(j)}, x_2^{(j)}, y_{\text{pref}}^{(j)}\}_{j=1}^{J_{t_0}}$, where $y_{\text{pref}}$ is $1$ if $x_1^{(j)}$ is preferred over $x_2^{(j)}$, and $0$ otherwise. Using $D_{\text{pref}}^{t_0}$, we can construct a binary preference function $g:\cX\times\cX\to\RR$ based on the following likelihood model for any candidate pair $x_1, x_2$:
\begin{equation*}
    \PP(y_{\text{pref}} \mid x_1, x_2) = S(y_\text{pref}; g(x_1, x_2)),
\end{equation*}
where $S(y_\text{pref}; z) := z^{y_\text{pref}} (1 - z)^{1 - y_\text{pref}}$ is the Bernoulli likelihood and $g(x_1, x_2)$ denotes the users degree of preference of $x_1$ over $x_2$.
There are various ways to learn such a $g$ \citep{bradley1952rank, chau2022spectral}. We choose to model $g$ with a GP to encapsulate the inherent estimation uncertainty, pivotal for designing an acquisition in $\S\ref{subsec:candidate_generation}$ that considers both surrogate and preference model's uncertainties. As the exact method we choose to learn $g$ is not the primary focus of this work, we defer this discussion in Supplementary~\ref{sup:pref}.

Figure \ref{fig:copeland} visualizes the GP preference function $g_{t_0}$ on the left, and $\hat{\pi}_{t_0}$ on the right, estimated by: %we can compute the global preference on $x$ as
\begin{align}
    \hat{\pi}_{g_{t_0}}(x) &:= \int_{\cX} g_{t_0}(x, x_2) d x_2.
    % \hat{\pi}_{g_t}(x) &:= \operatorname{Vol}_{\hat{\pi}_{g_{t_0}}}^{-1}\EE_{g_{t_0}}[\hat{\pi}_{g_{t_0}}(x)].
    \label{eq:pi_int}
\end{align}
This approach mirrors the soft-Copeland score\footnote{True Copeland score is calculated by Eq.(\ref{eq:pi_int}) but using true $\pi$ instead of estimated $\hat{\pi}$, thus there is no uncertainty.} in \citet{gonzalez2017preferential} and models the (unnormalized) likelihood of $x$ being the Condorcet winner\footnote{typically defined as the most favoured player within $\cX$.}. Hence, we can integrate out $g_{t_0}$ and obtain the following representation of our estimated user preference
\begin{equation*}
    \hat{\pi}_{{g_{t_0}}}(x) \sim \cN(\EE_{g_{t_0}}[\hat{\pi}_{g_{t_0}}(x)], \VV_{g_{t_0}}[\hat{\pi}_{g_{t_0}}(x)]).
\end{equation*}
Importantly, while the soft-Copeland score does not replicate the original function $f(x)$, the location of its maximum still corresponds to the maximum of $f(x)$. Furthermore, the maximum of the soft-Copeland score converges to the true maximum as the dataset size increases, namely $\lim_{t \rightarrow \infty} \mathop\mathrm{argmax}_{x \in \mathcal{X}} \pi_{g_t}(x) = \mathop\mathrm{argmax}_{x \in \mathcal{X}} f(x)$ if $|\mathcal{X}| < \infty$.
As the optimization progresses and the acquisition of more binary comparison data, we can iteratively update the posterior of $g_{t-1}$ via Bayes' theorem and recalibrate the preference model $\hat{\pi}_t$ at each iteration $t$.

\subsection{Candidate generation with no-harm guarantee}
\label{subsec:candidate_generation}
\begin{figure*}[hbt!]
  \centering
  \includegraphics[width=0.9\hsize]{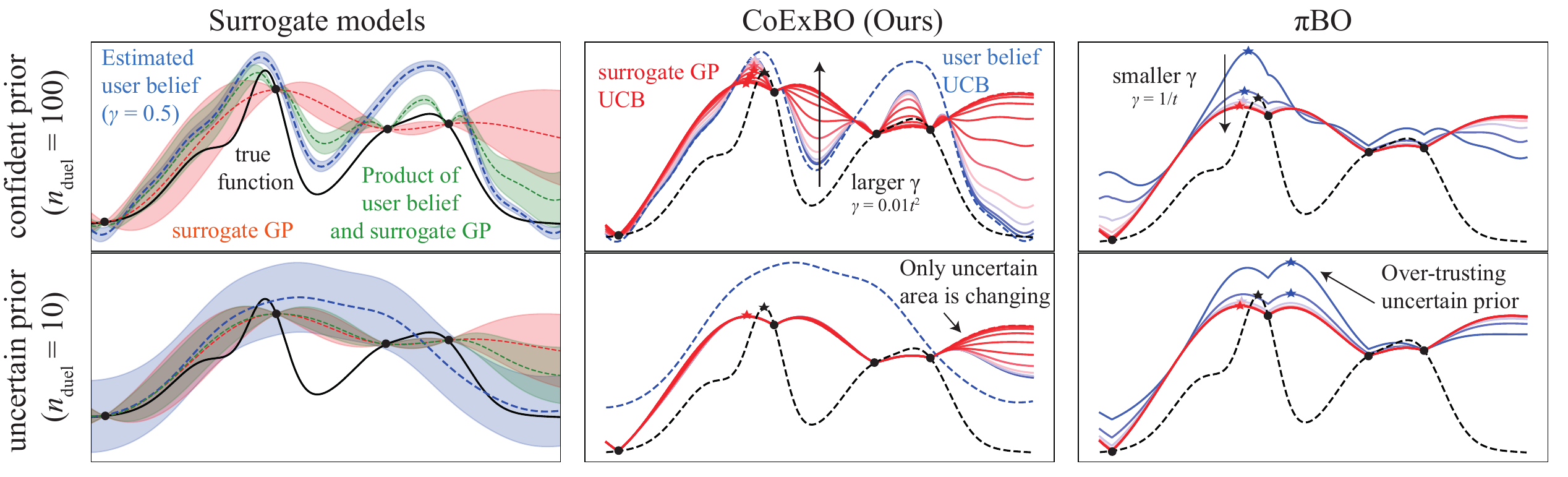}
  \caption{
  The CoExBO AF synthesizes GPs, utilizing one GP to represent the true function (red) and another to reflect user belief (blue), leading to the product GP (green). This product GP effectively assimilates the uncertainty inherent in user belief, adjusting the level of user belief integration during the acquisition process. Whereas the CoExBO AF is designed to adaptively manage the integration of uncertain user beliefs, the $\pi$BO approach tends to excessively depend on user belief, overlooking the uncertainty.%\Alan{We should describe the set up a bit clearer, this caption makes CoExBO sounds negative, why conservative is a good thing here?}
  }
  \label{fig:pibo}
  \vspace{-1.5em}
\end{figure*}
\textbf{New acquisition function. }
Following \citet{hvarfner2022pi}, we can use $\EE_{g_t}[\hat{\pi}_{g_t}]$ as a discounting factor for  $\alpha_{f_t}$ and redefine it as $\alpha_{f_t}(\cdot) \EE_{g_t}[\hat{\pi}_{g_t}(\cdot)]^{\frac{\gamma}{t}}$. However, this approach does not consider the predictive uncertainty of $\hat{\pi}_{g_t}$ represented by the GP model $g$, i.e., $\VV_{g_t}[\hat{\pi}_{g_t}(x)] = \EE_{g_t}[\hat{\pi}_{g_t}(x)(1-\hat{\pi}_{g_t}(x))]$. This could result in overly optimistic acquisitions.
$\pi$BO's performance depends on the peak centre of $\pi$, which must be close to the true global optimum for speedup. However, the peak centre of $\EE_{g_t}[\hat{\pi}_{g_t}]$ can be unreliable, especially when the user inputs are insufficient or inconsistent to confidently construct $\hat{\pi}_{g_t}$. Hence, we aim to utilize $\EE_{g_t}[\hat{\pi}_{g_t}]$ information only when it is confident, i.e., when $\VV_{g_t}[\hat{\pi}_{g_t}(x)]$ is sufficiently small.

To account for uncertainties in both the surrogate and preference model, we multiply the two Gaussians. For any $x\in\cX$, $f_t(x)$ is a Gaussian random variable in the target space of $f$, and $\hat{\pi}_{g_t}$ is a Gaussian random variable indicating the likelihood of $x$ being the Condorcet winner. We therefore scale the preference function to align with the surrogate model's scale. Using the property that the product of Gaussians is Gaussian, we derive a UCB-style AF.

\begin{proposition}\label{prop:acqf}
    Given $f_t(x)\sim \cN(\mu_{f_t}(x), \kappa_{f_t}(x,x))$, $\hat{\pi}_{g_t}(x)\sim \cN(\mu_{g_t}(x), \sigma_{g_t}^2(x))$ and a scaling function $\rho$ that maps $\hat{\pi}_{g_t}$ to the scale of $f_t$ and $\gamma > 0$, our new acquisition function $\alpha_{f, \pi}$ takes the following form:
    \begin{equation}
        \alpha_{f_t, \hat\pi_t}(x) := \mu_{f_t, \hat\pi_t}(x) + \beta^{\frac{1}{2}}\sigma_{f_t, \hat\pi_t}(x)
    \end{equation}
    where
    \begin{align}
         \mu_{f_t, \hat\pi_t} (x) &= \frac{\sigma^2_{f_t, \hat\pi_t} (x)}{\sigma^2_{\hat\pi_t} (x)} \mu_{\hat\pi_t} (x) + \frac{\sigma^2_{f_t, \hat\pi_t} (x)}{\sigma^2_{f_t} (x)} \mu_{f_t}(x),\\ \label{eq:post_mean}
         \sigma^2_{f_t, \hat\pi_t} (x) &= \frac{\sigma^2_{\hat\pi_t} (x) \sigma^2_{f_t} (x)}{\sigma^2_{\hat\pi_t}(x) + \sigma_{f_t}^2 (x)},\\
         \mu_{\hat\pi_t}(x) &:= \rho(\EE_{g_t}[\hat{\pi}_{g_t}(x)]),\\
         \sigma^2_{\hat\pi_t}(x) &:= \rho^2(\VV_{g_t}[\hat{\pi}_{g_t}(x)]) + \gamma t^2 \sigma^2_{f_t}(x),\label{eq:decay}\\
         \rho(x) &:= \EE[{\textbf{y}_\text{obj}}_t]x + \sqrt{\VV[{\textbf{y}_\text{obj}}_t]}
     \end{align}
    %\Alan{Masaki, please fill in the exact form of these $\mu_{f_t, \pi_t}$ and $\sigma_{f_t, \pi_t}$}
\end{proposition}
The new AF adheres to the following principle:
\begin{compactenum}
    \item (Uncertainty in Preference Estimation) The information provided by $\hat\pi_t(x)$ becomes valuable only when its variance, denoted as $\VV_{g_t}[\hat{\pi}_{g_t}(x)]$, is smaller than the variance of the surrogate model, represented as $\sigma_{f_t}^2(x)$.
    \item (Uncertainty in Efficacy of User Information) $\hat\pi_{g_t}(x)$ becomes less significant as iterations proceed. This mirrors $\pi$BO's principle that human knowledge is most valuable in the early stages.% of the process.
\end{compactenum}
The simple product of two Gaussian distributions offers a heuristic solution to the above assumptions. Firstly, the resulting Gaussian mean is a weighted sum of two means weighted by their variances, aligning with our first assumption, where the mean $\mu_{f_t, \hat\pi_t}$ stays unchanged from $\mu_{f_t}$ when $\VV_{g_t}[\hat{\pi}_{g_t}(x)] \gg \sigma_{f_t}^2(x)$. To address the second, we introduce a decay hyperparameter $\gamma$ term in Eq.~\ref{eq:decay}. When $\gamma t^2 \geq 1$, $\sigma^2_{\hat\pi_t}(x) > \sigma^2_{f_t}(x)$ holds, causing information from $\hat\pi_{g_t}$ to decay. While other methods could meet these principles, we choose the computationally simplest one.

Figure \ref{fig:pibo} illustrates typical behaviors of $\pi$BO and CoExBO. With a confident and accurate $\hat\pi_{g_t}$, both methods perform well. However, when dealing with uncertain $\hat\pi_{g_t}$, the peaks do not align with the true global maximum location. $\pi$BO tends to rely on user belief regardless of uncertainty, while CoExBO mitigates over-reliance on user belief in uncertain situations.

\textbf{Candidate generation.} Given $f_t$ and $\hat{\pi}_t$, we generate a pair of candidates as follows:
\begin{align}
    x_1 &= \argmax_{x\in \cX} \alpha_{f_t}(x) \tag{\small{standard UCB}} \\
    x_2 &= \argmax_{x\in \cX} \alpha_{f_t, \hat\pi_t}(x) \tag{\small{$\hat{\pi}$ incorporated UCB}}
\end{align}
This approach is similar to other human-AI collaborative BO methods \citep{gupta2023bo, khoshvishkaie2023cooperative}, involving a direct comparison of BO with human recommendations. Opting for $x_2$ speeds up convergence if human input is superior while choosing $x_1$ is optimal if BO performs better. This represents a greedy approach to optimizing choices for both sides. A greedy approach is optimal in both scenarios since we assume that either human or BO is superior.
The key difference lies in the decision-making process: in previous approaches, each agent independently selects their preferred option, necessitating separate queries. In contrast, our method employs BO to generate both candidates and then makes a selection. As our acquisition function $\alpha_{f, \hat\pi}$ gradually converges to the standard UCB, the selection process becomes equivalent over time. This prevents budget wastage resulting from suboptimal choices made by either side.

Note that our AF $\alpha_{f, \hat\pi}$ combines GP information and user beliefs, meaning it does not always align with user preferences. The GP component corrects any uncertainties or inaccuracies in user beliefs, preventing a persistent bias toward selecting $x_2$.

\textbf{Regret analysis.}
%\Alan{Please go through your writing again and align notations as above.}
By following \citet{srinivas2009gaussian}, we analyse the regret of preference-based AF $r_{\hat\pi_t} := f(x^*_\text{true}) - f(x_2)$ and the standrd UCB regret $r_t := f(x^*_\text{true}) - f(x_1)$. We assume good and bad user beliefs. A good user belief assumes to contain the true function within the standard deviation, whereas a bad user belief does with extra error with mean estimation.
\begin{theorem}\label{thm:regret}
    Fix $t \geq 1$ and $\gamma > 0$. If $\lvert f(\boldsymbol{x}) - \mu_{f_{t-1}}(\boldsymbol{x}) \rvert \leq \beta_t^{1/2} \sigma_{f_{t-1}}(\boldsymbol{x})$ for all $\boldsymbol{x} \in D$ , $\lvert D \rvert < \infty$ hold, the ratio of regrets for with and without $\pi$ augmentation $r_t, r_{\hat\pi_t} $ is bounded by:\\
    (Good user belief) If $\lvert f(\boldsymbol{x}) - \mu_{f_{t-1}, \hat\pi_{t-1}}(\boldsymbol{x}) \rvert \leq \beta_t^{1/2} \sigma_{f_{t-1}, \hat\pi_{t-1}}(\boldsymbol{x})$ holds,
    \begin{align}
        r_{\hat\pi_t} {r_t}^{-1} \leq {R_{\hat\pi_t}} < 1,
    \end{align}
    where\\
    $R_{\hat\pi_t} = \sqrt{\frac{\rho^2(\mathbb{V}_{g_{t-1}}[\hat\pi_{g_{t-1}}(x_2)]) + \gamma (t -1)^2 \sigma^2_{t-1}(x_2)}{\rho^2(\mathbb{V}_{g_{t-1}}[\hat\pi_{g_{t-1}}(x_2)]) + \gamma (t-1)^2 \sigma^2_{t-1}(x_2) + \sigma^2_{t-1}(x_1)}}$.\\
    (Bad user belief)
    If $\lvert f(\boldsymbol{x}) - \mu_{f_{t-1}, \hat\pi_{t-1}}(\boldsymbol{x}) \rvert \leq \lvert \mu_{f_{t-1}}(\boldsymbol{x}) -  \mu_{f_{t-1}, \hat\pi_{t-1}}(\boldsymbol{x}) \rvert + \beta_t^{1/2} \sigma_{f_{t-1}, \hat\pi_{t-1}}(\boldsymbol{x})$ holds,
    \begin{align}
        r_{\hat\pi_t} {r_t}^{-1} \leq \Delta \mu_t + {R_{\hat\pi_t}}_t,
    \end{align}
    where $\Delta \mu_t = \frac{\lvert \mu_{t-1}(x_1) -  \mu_{f_{t-1}, \hat\pi_{t-1}}(x_2) \rvert}{2 \beta_t^{1/2} \sigma_{f_{t-1}}(x_1)}$.\\
\end{theorem}
The proof is given in Supplementary \ref{sup:proof}. Using Theorem \ref{thm:regret}, we obtain the convergence rate of $\alpha_{f_{t-1}, \hat\pi_{t-1}}$. This trivially follows the original convergence rate on UCB as in \citet{srinivas2009gaussian}:
\begin{lemma}\label{thm:convergence}
    (No harm guarantee) Given the regret in Theorem \ref{thm:regret}, The regret of a preference-based acquisition function, ${\alpha_\text{pref}}_t$, asymptotically equals to the regret of an upper confidence bound strategy, UCB:
    \begin{align}
        \lim_{t \rightarrow \infty}  r_t^\pi {r_t}^{-1} = 1,
    \end{align}
    so we obtain a convergence rate for $\alpha_{f_T, \hat\pi_T}$ of $\mathcal{O}(\sqrt{T \gamma_T \log T})$, an original UCB convergence rate.
\end{lemma}
Hence, we can ensure that the worst-case convergence rate remains unaffected, even with inaccurate user beliefs, for large $t$, where $\gamma t^2 \gg 1$. While short-term performance may not match the standard UCB, it often yields better empirical results. Particularly, a good user belief has a provably better regret bound. In our scenario, humans choose candidates from the UCB or $\alpha_{f_t, \hat\pi_t}$, reinforcing the no-harm guarantee. The human selection process does not impact the convergence rate because $\alpha_{f_t, \hat\pi_t}$ determines tighter or looser bounds than the UCB, depending on user beliefs. In practice, human knowledge evolves over iterations, positively influencing convergence, as demonstrated in our experiments section. The parameter $\gamma t^2$ balances the integration of evolving human knowledge with the no-harm guarantee. It is worth noting that our approach differs from that of multitask GPs, as multitask GPs are vulnerable to unreliable low-fidelity GPs \citep{mikkola2023multi}.

\subsection{Explaining candidate generation through Shapley values}\label{subsec:explanation}

\begin{table*}[htp]
    \centering
    \caption{Comparisons between our proposed CoExBO with baselines used in the ablation study.}
    \resizebox{0.75\textwidth}{!}{
    \begin{tabular}{lllllll}
    \toprule
    &&&&\multicolumn{3}{c}{Novel contributions}\\
    \cmidrule{5-7}
         Baselines &
         \begin{tabular}{@{}l@{}}Human \\ selection\end{tabular}&
         BO &
         \begin{tabular}{@{}l@{}}$\pi$ aug- \\ mentation\end{tabular}&
         \begin{tabular}{@{}l@{}}Interative \\ $\pi$ update\end{tabular}&
         \begin{tabular}{@{}l@{}}Uncertainty \\ in $\pi$ estimation\end{tabular}&
         Explanation\\
    \midrule
         random& \ding{51} & \xmark & \xmark& \xmark& \xmark& \xmark\\
         manual search& \ding{51} &\xmark &\xmark  &\xmark  &\xmark  &\xmark \\ 
         UCB/TS&  \xmark & \ding{51} &\xmark  &\xmark  &\xmark  &\xmark \\ 
         prior sampling& \xmark & \xmark &  \ding{51}&\xmark  &\xmark  &\xmark \\ 
         batch UCB/TS& \ding{51} & \ding{51}& \xmark & \xmark & \xmark & \xmark\\ 
         $\pi$BO \citep{hvarfner2022pi} &  \xmark & \ding{51}  &  \ding{51}& \xmark & \xmark & \xmark\\ 
         CoExBO ($\pi$BO)&  \ding{51}&  \ding{51}&  \ding{51}&  \ding{51}& \xmark &\xmark \\ 
         CoExBO&  \ding{51}&  \ding{51}&  \ding{51}&  \ding{51}&  \ding{51}&  \ding{51}\\
    \bottomrule
    \end{tabular}
    }
    \label{tab:comparison}
\end{table*}

To foster trust in the black-box optimizer among users, we employ Shapley values~\citep{shapley1953value}, a popular solution concept from game theory adopted by the machine learning community~\citep{lundberg2017unified, chau2022rkhs, hu2022explaining} to provide feature attributions for the acquisitions and the surrogate model. This provides users with a clearer understanding of the factors influencing the selection of candidates.

Shapley values follow a set of favourable rationality axioms, setting them apart from heuristic methods like extracting the length scale from a GP kernel. For a given function $h:\cX\to\RR$, a query location $x$, the Shapley value for feature $j$ is expressed as 
\begin{align*}
    \phi_{j, x}(h) = &\sum_{S\subseteq [d]\backslash \{j\}} c_{|S|} \left(\nu_{x, h}(S\cup i) - \nu_{x, h}(S)\right)
\end{align*}
where $[d]:=\{1,\dots, d\}$, $c_{|S|} = \frac{1}{d}{\binom{d-1}{|S|}}^{-1}$, $X_S$ is the subfeature vector of $X$ for features in $S$ and $\nu_{x, h}(S)$ measures a notion of contribution features $S$ has to the prediction $h(x)$. 
We utilize the recently introduced GPSHAP~\citep{chau2023explaining} to explain the surrogate GP $f$. We illustrate how to estimate the Shapley values for $\alpha_f$, but extending to $\alpha_{f, \pi}$ is straightforward. While in the Shapley explanation literature, it is suggested one should take the conditional expectation of the to-be-explained function, i.e. $\EE_X[\alpha_{f}(X)\mid X_S=x_s]$, to structure the cooperative game. However, for computational reasons (see appendix), we opted to establish the game using its upper bound instead:
\begin{align*}
    \nu_{x, f}(S)&:=\EE[\mu_f(X) \mid X_S=x_S] \\
    &\quad \, + \beta_t^{1/2} \sqrt{\EE[\kappa_f(X, X)  \mid X_S=x_S]}.
\end{align*}
$\nu_{x, f}(S)$ can be interpreted as the significance of the feature subset $S$ measured by how much the upper bound has changed if we remove the contribution from other features in $S^c$ by integration. This formulation allows us to estimate the quantity in analytical form:
\begin{proposition}
    Given $f\sim\cG\cP(\mu, \kappa)$, for a given feature subset $S\subseteq \{1,\dots, d\}$, and location $x, \nu_{x, f}(S)$ can be estimated from observations as
    \begin{align}
        \bfB_{S}(\bfx)^\top \tilde{\bff} + \beta_t^{1/2} \sqrt{\bfB_S(\bfx)^\top\tilde{\bfK}_{\bfX\bfX}\bfB_S(\bfx)}
    \end{align}
    where $\bfB_S(\bfx) = (\bfK_{\bfX_S\bfX_S} + \lambda_S I)^{-1}k_S(\bfX_S,\bfx_S)$, $\lambda_S>0$ is a regularisation parameter, $\tilde{\bff}$ is the posterior mean, and $\tilde{\bfK}_{\bfX, \bfX}$ is the posterior covariance matrix of the GP.
\end{proposition}
To obtain this quantity, we utilized the fact that the conditional expectation of GPs also admits an analytical form; see \citet{chau2021deconditional,chau2021bayesimp} for further details.
Other explanation features based on Shapley values are detailed in Supplementary~\ref{sup:explain}.

% \vspace{-1em}
\section{Experiments}\label{sec:experiments}
\begin{figure*}
  \centering
  \includegraphics[width=0.8\hsize]{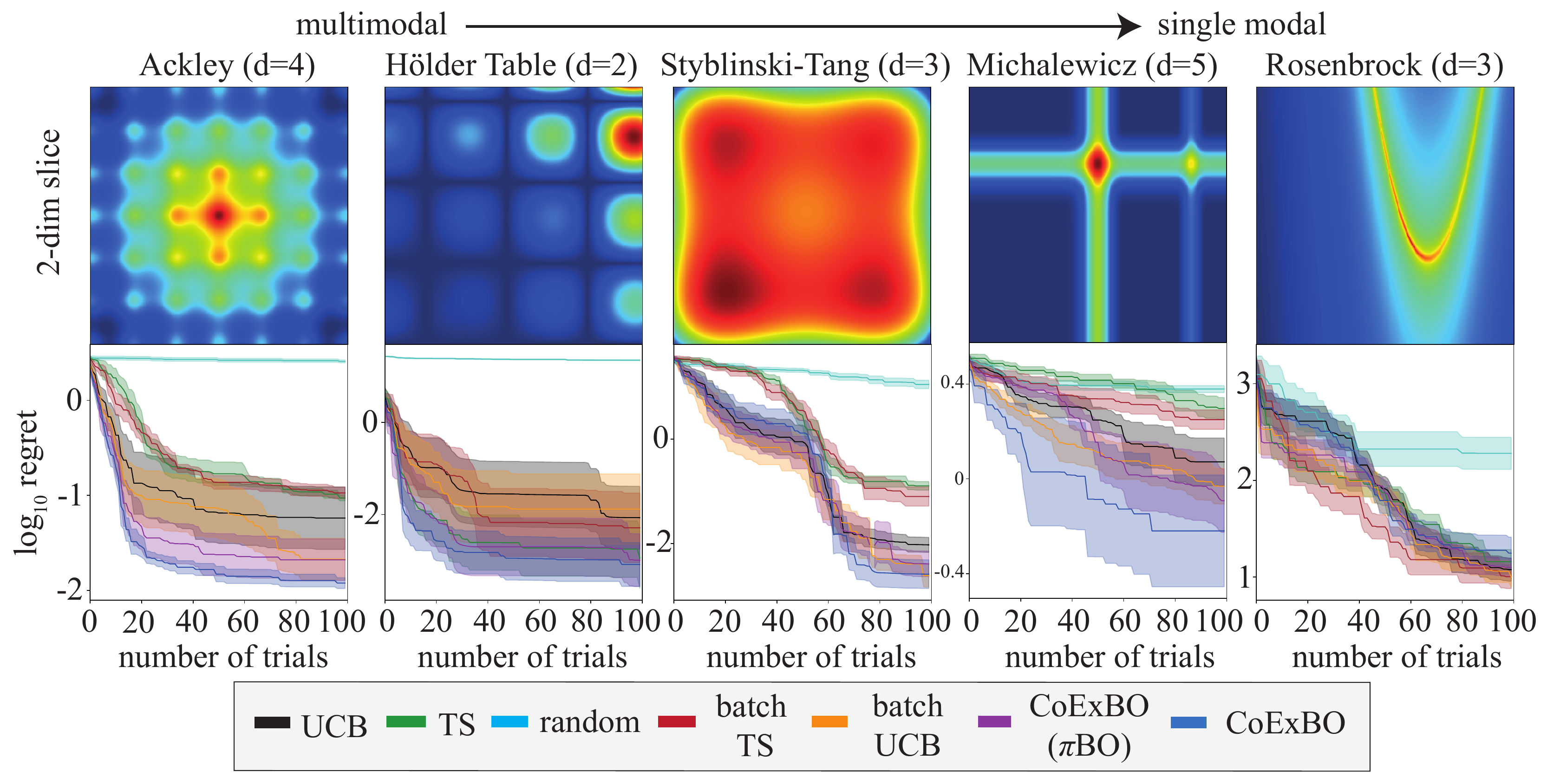}
  \caption{Convergence plot of simple regret for 5 synthetic functions with the synthetic selection accuracy ($\epsilon_\text{pref} := \mathcal{N}(0, 0.1^2)$). Lines and shaded area denote mean $\pm$ 1 standard error. CoExBO consistently outperforms all six baselines except for the Rosenbrock function. The dark red region is the global maximum.}
  \label{fig:ackley}
  \vspace{-1em}
\end{figure*}
CoExBO has been tested for synthetic and real-world tasks and is implemented using PyTorch \citep{paszke2019pytorch}, GPyTorch \citep{gardner2018gpytorch}, BoTorch \citep{balandat2020botorch}, and SOBER \citep{adachi2023sober}. All experiments were averaged over 10 repeats, computed with a laptop PC\footnote{MacBook Pro 2019, 2.4 GHz 8-Core Intel Core i9, 64 GB 2667 MHz DDR4}. We set the initial random samples for objective queries as $n_\text{obj} = 10$ and for preferential learning $n_\text{pref} = 100$, respectively.

Table \ref{tab:comparison} provides a summary of the baseline algorithms we evaluated. These include \textbf{Random}: This method generates a pair of i.i.d. samples uniformly, after which a human selects the preferred one. \textbf{Manual Search}: In this approach, a human selects the next query without any algorithmic assistance. \textbf{UCB} \citep{srinivas2009gaussian} and \textbf{Thompson Sampling (TS)} \citep{thompson1933likelihood}: Both methods autonomously select the next query without human intervention. \textbf{Prior Sampling}: This technique involves choosing the next query point as an i.i.d. sample from the estimated prior $\hat{\pi}$, derived from the initial preference samples $n_\text{pref}$, without BO assistance. \textbf{BatchUCB} \citep{azimi2010batch} and \textbf{BatchTS} \citep{kandasamy2018parallelised}: These algorithms generate pairs of candidates, from which a human selects one. They do not integrate human knowledge $\hat{\pi}$ in the candidate generation process. \textbf{$\pi$BO} \citep{hvarfner2022pi}: This algorithm selects the next query point based on a $\hat{\pi}$-augmented AF, incorporating human knowledge through $\hat{\pi}$. However, it is not interactive (as it is fixed before running the BO) and does not account for uncertainty in $\hat{\pi}$ estimation or human interactive selection. Our proposed algorithm, \textbf{CoExBO}, incorporates all these elements. Its variant, \textbf{CoExBO ($\pi$BO)}, specifically analyzes the efficacy of our new AF by replacing it with the $\pi$BO's AF, $\alpha_{f_t}(\cdot) \EE_{g_t}[\hat{\pi}_{g_t}(\cdot)]^{\frac{\gamma}{t}}$. Following the methodology in the original $\pi$BO paper, we set the decaying hyperparameter to 10 for $\pi$BO and $\gamma$ to 0.01 for our algorithm.

\subsection{Synthetic Functions with Synthetic Human Selection}
\begin{figure}
  \centering
  \includegraphics[width=\hsize]{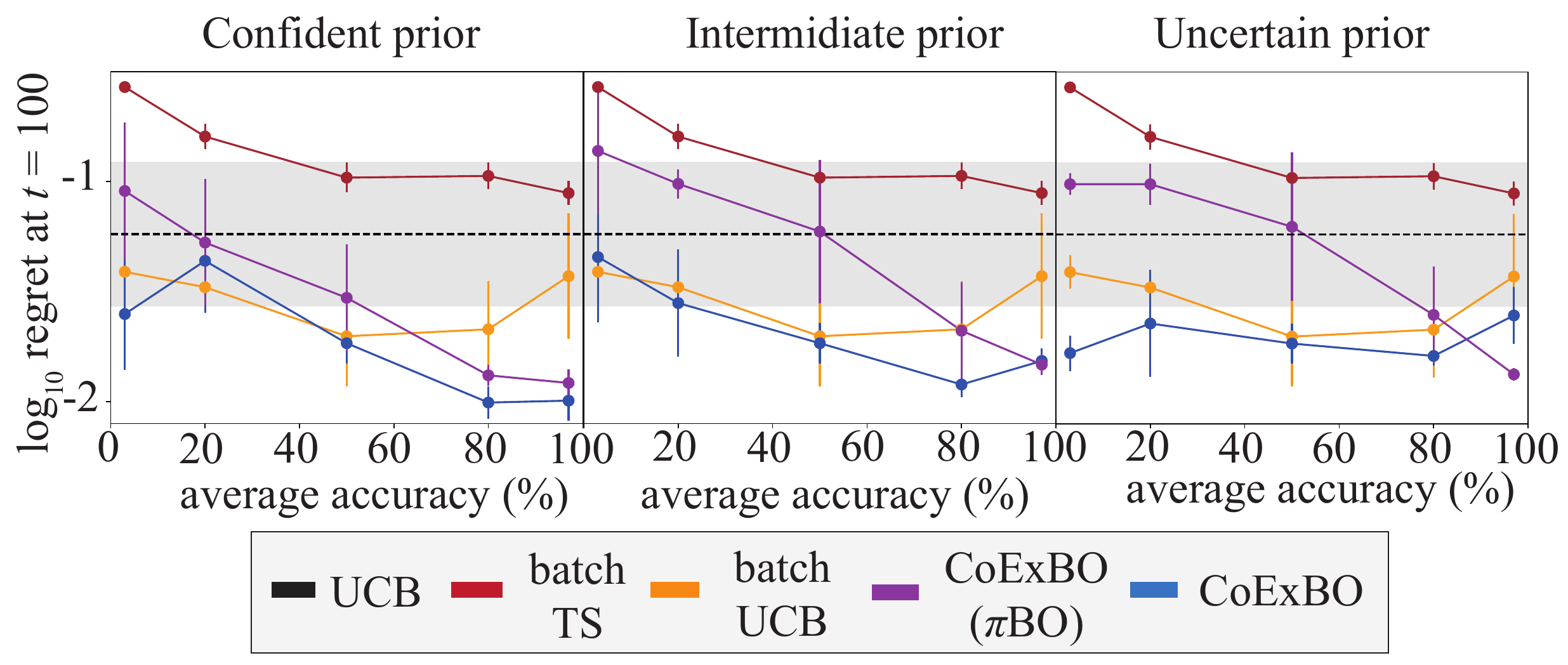}
  \caption{Convergence after 100 iterations on Ackley function ($d=4$) with three prior confidence levels and five selection accuracy levels. Lines and error bars denote mean $\pm$ 1 standard error.}
  \label{fig:robustness}
  \vspace{-1.5em}
\end{figure}
\begin{figure*}
  \centering
  \includegraphics[width=0.7\textwidth]{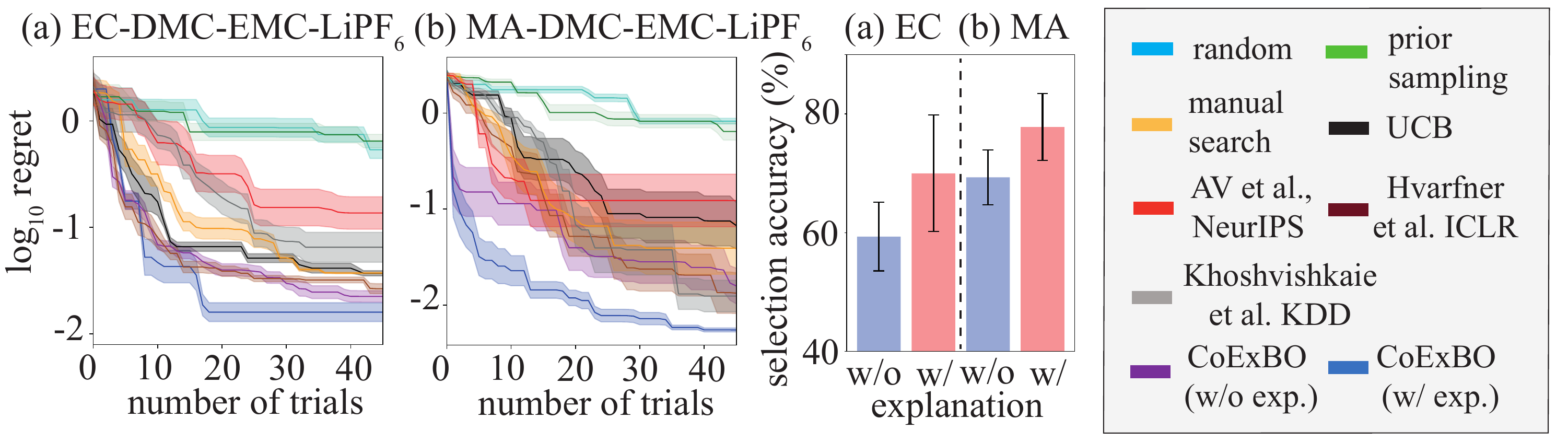}
  \caption{Convergence plot and selection accuracy of two battery material design tasks (a) EC (b) MA-based system. The explainability strengthens the selection accuracy and accelerates convergence.}
  \label{fig:electrolyte}
  \vspace{-1.5em}
\end{figure*}
\begin{figure}
  \centering
  \includegraphics[width=0.7\hsize]{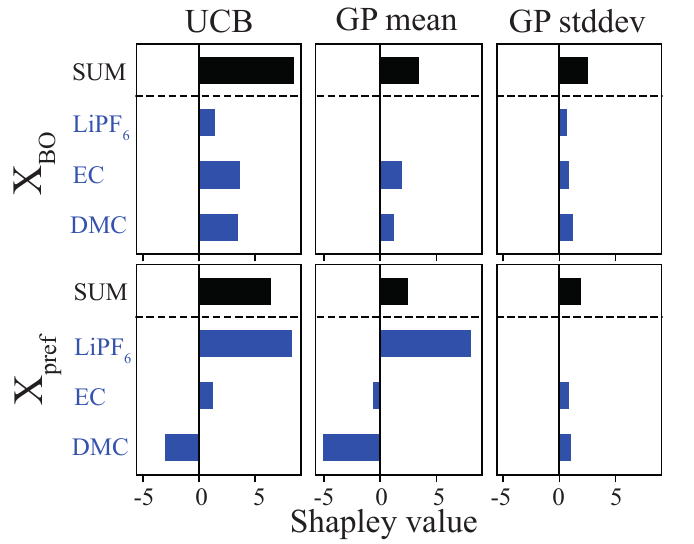}
  \caption{An example of Shapley values for UCB, GP predictive mean and standard deviation during the iteration at two recommended locations.}
  \label{fig:explain}
  \vspace{-1.5em}
\end{figure}

\textbf{Synthetic functions.} First, CoExBO was tested with synthetic functions and a synthetic human selection as $\operatorname{H}(x_1,x_2)$ such that $f_\text{human}(x_1) > f_\text{human}(x_2)$, where $f_\text{human}(x) := f(x) + \epsilon_\text{pref}$, and $\epsilon_\text{pref} \sim \mathcal{N}(x; 0, \sigma^2_\text{pref})$. The correct human selection rate can be modified by changing $\sigma^2_\text{pref}$. $\sigma^2_\text{pref} = 0.1$ throughout this experiment. We have chosen the five commonly used test functions \citep{simulationlib} (see details in Supplementary~\ref{sup:synthetic}). 

Figure \ref{fig:ackley} shows the results on simple regret. CoExBO consistently outperforms four baselines except for the Rosenbrock function. This suggests human feedback is particularly effective for multimodal functions with many local minima. This is evident as there are no big differences between the algorithms with and without human intervention in the Rosenbrock function. Further details and computational complexity analysis can be found in Supplementary~\ref{sup:synthetic}.

\textbf{Robustness evaluation.}
We tested CoExBO's robustness with the Ackley function regarding (a) uncertain prior and (b) incorrect human selections. We varied the prior confidence by changing the number of initial random samples ($n_\text{pref} = 10, 100, 500$). To vary human selection correctness, we adjusted the noise variance of the synthetic human function ($\sigma_\text{pref}^2 = 0.1, 1, 100$). We simulated adversarial selection cases by flipping the feedback from $\sigma_\text{pref}^2 = 0.1, 1$.

Figure \ref{fig:robustness} demonstrates that CoExBO is robust against uncertain prior knowledge and incorrect human selections. While the optimistic $\pi$BO AF becomes less effective with reduced selection accuracy, CoExBO maintains its effectiveness better. The key distinction between $\pi$BO and CoExBO AFs is that the former only modifies the UCB in uncertain areas, whereas $\pi$BO adjusts it regardless of uncertainty (see Figure \ref{fig:pibo}). In other words, once the wrongly believed position is queried, the uncertainty in that area decreases, leading to unbiased UCB. This feature offers greater resilience to incorrect and uncertain human selections compared to $\pi$BO AF. Note that both $\pi$BO and CoExBO are guaranteed to asymptotically approach the standard UCB, so longer iterations should yield similar results.
Both batchUCB and CoExBO exhibited robustness against adversarial selection. However, the difference in adversarial selection falls within the standard error of UCB, indicating that it does not outperform the standard UCB in adversarial cases. Relative differences in tendencies provide more reliable insights. Further details are explained in Supplementary \ref{sup:synthetic}.

\subsection{Real-World Tasks with Human Experts}
Lithium-ion batteries are the key to realizing the electrification of many sectors as a climate action. However, due to their complicated chemical nature, a perfect simulator that predicts required material properties under all operational and degradation conditions does not exist. Hence, researchers need to repeat costly laboratory experiments to find the best combinations of materials from which to build new lithium-ion batteries.

We assessed expert advice's effectiveness with four battery researchers and compared results with and without explainability features to gauge their impact on selection accuracy. The problem involves finding the best electrolyte material combination to maximize ionic conductivity. This task is challenging due to complex solvation and many-body effects \citep{gering2017prediction}, making prediction with a simulator-based approach difficult. We applied CoExBO to two electrolyte design problems: one involving four materials (EC-DMC-EMC-LiPF$_6$) \citep{dave2022autonomous}, a well-known combination, and the other comprising MA-DMC-EMC-LiPF$_6$ \citep{logan2018study}, an unfamiliar composition to all participating experts. While materials science knowledge can deduce the effect on lithium-ion solvation states by changing from carbonate to acetate non-aqueous solvents, their knowledge is qualitative and not quantitative.

Figure \ref{fig:electrolyte} shows that CoExBO with expert knowledge accelerates convergence, even without explainability features. The addition of explainability results in a significant speedup in time-to-accuracy and enhancement in selection accuracy, which outperformed eight baselines. Figure \ref{fig:explain} exemplifies a typical case of Shapley values. The black bar's sum of Shapley values shows that $X_\text{BO}$ has a higher UCB value, indicating that selecting $X_\text{BO}$ is natural in standard BO. However, the preference-based $X_\text{pref}$ attributes more reasonable importance to conductivity, consistent with chemical expertise—highlighting LiPF$_6$ as the key material. %These findings illustrate our algorithm's ability to adaptively transfer implicit and qualitative expert knowledge to an unknown system and the efficacy of explainability.
We illustrate the CoExBO's effectiveness with two distinct cases:
One participant initially relied heavily on BO suggestions, a phenomenon known as automation bias \citep{cummings2004automation}. \citet{goddard2012automation, skitka2000accountability} have shown that explaining the process and holding users accountable for their decision accuracy can reduce such biases, both of which CoExBO achieved.
Another participant consistently trusted their own preferences, even when the GP improved. CoExBO's no-harm guarantee shepherds users to global convergence.
%, but the explainability feature had only a minor impact on improving selection accuracy.

In summary, expert knowledge can be particularly helpful to the GP in two ways: (A) Assessing the reliability of noisy ionic conductivity measurements and disregarding noise to infer the true function shape. (B) Applying their chemical knowledge to roughly optimize the composition, exploiting the knowledge of each material's importance. On the other hand, CoExBO can help experts in three ways: (A) BO is better at fine-tuning more precisely than experts, as expert knowledge only focuses on the main effect. (B) Shapley values provide accurate importance rankings conditioned on $x$, which updates experts' knowledge. (C) The feedback and explaining feature can correct experts' wrong understanding and guide them in the true optimal direction.

% \vspace{-1em}
\section{Discussions and Limitations}
Our approach accelerates convergence when experts possess accurate comparative knowledge that the GP cannot access, which remains a strong assumption, albeit weaker than the conventional ones with a well-defined and error-free belief function or the optimal query. Expert knowledge can be particularly helpful to the GP in three ways: (a) as a good warm starter, allowing it to begin from a promising region; (b) as a good encoder, compensating for the information lost during simplification to a low-dimensional search space; and (c) as a noise reducer, providing a more accurate estimation of experimental noise. These aspects align well with fields such as chemistry and scientific experiments, and experts can convey this complex information through simple selections.
Our proofs are specific to the UCB setting, but the decay property can ensure convergence for any AF. %, making the theoretical guarantee useful for various applications. 
%High-dimensional space presents a shared challenge with BO and preferential learning, yet recent advancements may help overcome this limitation. 
In our experiments with experts, we observed that there is a tendency to expect both surrogate and explanation models to provide an `oracle' understanding of the whole scientific process. We emphasise this is not the purpose of explainable BO as we are by definition, operating under a small data regime. However, our collaboration and explanation framework allows us to demystify the BO process and thus mitigate over-trust.

% Some participants had unrealistically high expectations, hoping for `oracle' explanations to discover the true global optimum. However, our explainability demystifies the BO processes and mitigates over-trust.

There is a growing interest in putting humans back into the optimization cycle. A prime example is the RLHF to fine-tune LLMs \citep{christiano2017deep, rafailov2024direct}. We also observe concurrent works centered on enhancing human-AI collaboration \citep{av2024enhanced} and explainable BO \citep{chakraborty2023post, chakraborty2024explainable}, showcasing this as a promising direction of research.
%\Alan{We can mention the increasing interests in putting human back in the loop and draw some parallel to RLHF, also mention concurrent works here, say something by the time this work has been accepted there are already a few concurrent work etc.}

\newpage
\subsubsection*{Acknowledgements}
We thank Philipp Dechent and Katie Lukow for participating in the real-world experiments, and anonymous reviewers who gave useful comments. Masaki Adachi was supported by the Clarendon Fund, the Oxford Kobe Scholarship, the Watanabe Foundation, and Toyota Motor Corporation.

\bibliography{aistats2024_conference}
\bibliographystyle{aistats2024_conference}

 \begin{enumerate}
 \item For all models and algorithms presented, check if you include:
 \begin{enumerate}
   \item A clear description of the mathematical setting, assumptions, algorithm, and/or model. [Yes]
   \item An analysis of the properties and complexity (time, space, sample size) of any algorithm. [Yes in supplementary]
   \item (Optional) Anonymized source code, with specification of all dependencies, including external libraries. [Yes at https://anonymous.4open.science/r/CoExBO-4B06/]
 \end{enumerate}

 \item For any theoretical claim, check if you include:
 \begin{enumerate}
   \item Statements of the full set of assumptions of all theoretical results. [Yes]
   \item Complete proofs of all theoretical results. [Yes]
   \item Clear explanations of any assumptions. [Yes]     
 \end{enumerate}

 \item For all figures and tables that present empirical results, check if you include:
 \begin{enumerate}
   \item The code, data, and instructions needed to reproduce the main experimental results (either in the supplemental material or as a URL). [Yes/No/Not Applicable]
   \item All the training details (e.g., data splits, hyperparameters, how they were chosen). [Yes in text and supplementary]
         \item A clear definition of the specific measure or statistics and error bars (e.g., with respect to the random seed after running experiments multiple times). [Yes]
         \item A description of the computing infrastructure used. (e.g., type of GPUs, internal cluster, or cloud provider). [Yes]
 \end{enumerate}

 \item If you are using existing assets (e.g., code, data, models) or curating/releasing new assets, check if you include:
 \begin{enumerate}
   \item Citations of the creator If your work uses existing assets. [Yes]
   \item The license information of the assets, if applicable. [Not Applicable]
   \item New assets either in the supplemental material or as a URL, if applicable. [Not Applicable]
   \item Information about consent from data providers/curators. [Yes in supplementary]
   \item Discussion of sensible content if applicable, e.g., personally identifiable information or offensive content. [Not Applicable]
 \end{enumerate}

 \item If you used crowdsourcing or conducted research with human subjects, check if you include:
 \begin{enumerate}
   \item The full text of instructions given to participants and screenshots. [Yes]
   \item Descriptions of potential participant risks, with links to Institutional Review Board (IRB) approvals if applicable. [Not Applicable]
   \item The estimated hourly wage paid to participants and the total amount spent on participant compensation. [Not Applicable]
 \end{enumerate}

 \end{enumerate}

\appendix
\newpage
\onecolumn
\addcontentsline{toc}{section}{Appendix} % Add the appendix text to the document TOC
\part{Appendix} % Start the appendix part
\parttoc % Insert the appendix TOC

\section{Proof of theorem} \label{sup:proof}
\subsection{Regret analysis of normal UCB policy}
We begin with the finite case, $|D| < \infty$.
We recall two lemmas from \citet{srinivas2009gaussian}.
\begin{lemma}
    Pick $\delta \in (0,1)$ and set $\beta_t = 2 \log(|D| p_t / \delta)$, where $\sum_{t \geq 1} p_t^{-1} = 1$, $p_t > 0$. Then,
    \begin{align}
        \lvert f(\boldsymbol{x}) - \mu_{f_{t-1}}(\boldsymbol{x})) \rvert \leq \beta_t^{1/2} \sigma_{f_{t-1}} (\boldsymbol{x}) \quad \forall \boldsymbol{x} \in D, \, \forall t \geq 1
    \end{align}
    holds with probability $\geq 1 - \delta$. 
\end{lemma}

\textbf{Proof} Fix $t \geq 1$ and $\boldsymbol{x} \in D$. Conditioned on $\boldsymbol{y}_{t-1} = (y_1, \dotsc, y_{t-1})$, $\{\boldsymbol{x}_1,\dotsc, \boldsymbol{x}_{t-1}\}$ are deterministic, and $f(\boldsymbol{x}) \sim \mathcal{N} (\mu_{f_{t-1}}(\boldsymbol{x}), \sigma^2_{f_{t-1}}(\boldsymbol{x})$. Now, if $r \sim \mathcal{N}(0,1)$, then
\begin{align}
    \mathbb{P}(r > c) &= \exp \left( -\frac{c^2}{2} \right) (2p)^{-1/2} \int \exp \left( -\frac{(r - c)^2}{2} - c(r - c) \right) \text{d} r,\\
    &\leq \exp \left( -\frac{c^2}{2} \right) \mathbb{P}(r > 0),\\
    &= \frac{1}{2} \exp \left( -\frac{c^2}{2} \right)
\end{align}
for $c > 0$, since $\exp \left(- c(r - c) \right) \leq 1$ for $r > c$. Therefore, $\mathbb{P}\left( \lvert f(\boldsymbol{x}) - \mu_{f_{t-1}}(\boldsymbol{x})) \rvert \leq \beta_t^{1/2} \sigma_{f_{t-1}} (\boldsymbol{x}) \right) \leq \exp  \left(- \frac{\beta_t}{2} \right)$, using $r = \frac{f(\boldsymbol{x}) - \mu_{f_{t-1}}(\boldsymbol{x}))}{\sigma_{f_{t-1}} (\boldsymbol{x})}$ and $c = \beta_t^{1/2}$. Applying the union bound,
\begin{align}
    \lvert f(\boldsymbol{x}) - \mu_{f_{t-1}}(\boldsymbol{x})) \rvert \leq \beta_t^{1/2} \sigma_{f_{t-1}} (\boldsymbol{x}) \quad \forall \boldsymbol{x} \in D
\end{align}
holds with probability $\geq 1 - |D| \exp  \left(- \frac{\beta_t}{2} \right)$. Choosing $|D| \exp  \left(- \frac{\beta_t}{2} \right) = \frac{\delta}{p_t}$ and using the union bound for $t \in \mathbb{N}$, the statement holds. 

\begin{lemma}
    Fix $t \geq 1$. If $\lvert f(\boldsymbol{x}) - \mu_{f_{t-1}}(\boldsymbol{x})) \rvert \leq \beta_t^{1/2} \sigma_{f_{t-1}} (\boldsymbol{x})$ for all $\boldsymbol{x} \in D$, then the regret $r_t$ is bounded by $2 \beta^{1/2}_t \sigma_{f_{t-1}}(\boldsymbol{x})$. 
\end{lemma}

\textbf{Proof} By definition of $\boldsymbol{x}_\text{bo} := \mathop\mathrm{argmax}_{x \in \mathcal{X}} \mu_{f_{t-1}}(\boldsymbol{x}) + \beta^{1/2}_t \sigma_{f_{t-1}}(\boldsymbol{x})$, we have  
$\mu_{f_{t-1}}(\boldsymbol{x}_\text{bo}) + \beta^{1/2}_t \sigma_{f_{t-1}}(\boldsymbol{x}_\text{bo}) \geq \mu_{f_{t-1}}(\boldsymbol{x}^*_\text{true}) + \beta^{1/2}_t \sigma_{f_{t-1}}(\boldsymbol{x}^*_\text{true}) \geq f_\text{true}(\boldsymbol{x}^*_\text{true})$. Therefore

\begin{align}
    r_t = f(\boldsymbol{x}^*_\text{true}) - f(\boldsymbol{x}_\text{bo})
    &\leq \beta^{1/2}_t \sigma_{f_{t-1}}(\boldsymbol{x}_\text{bo}) + \mu_{f_{t-1}}(\boldsymbol{x}_\text{bo}) - f(\boldsymbol{x}_\text{bo})\\
    &\leq 2\beta^{1/2}_t \sigma_{f_{t-1}}(\boldsymbol{x}_\text{bo})
\end{align}

\subsection{Proof of Regrets}
Recall the definition of the good user belief is $\lvert f(\boldsymbol{x}) - \mu_{f_{t-1}, \hat{\pi}_{t-1}}(\boldsymbol{x})) \rvert \leq \beta_t^{1/2} \sigma_{f_{t-1}, \hat{\pi}_{t-1}} (\boldsymbol{x})$.
\paragraph{Proof of good user belief regrets}
\begin{align}
    \frac{r_{\hat\pi_t}}{r_t} 
    &= \frac{f(x^*_\text{true}) - f(x_2)}{f(x^*_\text{true}) - f(x_1)}\\
    &\leq \frac{2\beta^{1/2}_t \sigma_{f_{t-1}, \hat{\pi}_{t-1}}(x_2)}{2\beta^{1/2}_t \sigma_{f_{t-1}}(x_1)} &&\text{(Lemma 5)}\\
    &= \frac{\sigma_{f_{t-1}, \hat{\pi}_{t-1}}(x_2)}{\sigma_{f_{t-1}}(x_1)}\\
    &= \frac{\sigma_{\hat{\pi}_{t-1}}(x_2)}{\sqrt{\sigma_{\hat{\pi}_{t-1}}^2(x_2) + \sigma^2_{f_{t-1}}(x_1)}} &&\text{(Proposition 1)}\\
    &= \sqrt{\frac{\rho^2(\mathbb{V}_{g_{t-1}}[\hat\pi_{g_{t-1}}(x_2)]) + \gamma (t-1)^2 \sigma^2_{f_{t-1}}(x_2)}{\rho^2(\mathbb{V}_{g_{t-1}}[\hat\pi_{g_{t-1}}(x_2)]) + \gamma (t-1)^2 \sigma^2_{f_{t-1}}(x_2) + \sigma^2_{f_{t-1}}(x_1)}} &&\text{(Proposition 1)}\\
    &= R_{\hat\pi_t}\\
    &< 1 &&(\sigma^2_{f_{t-1}}(x_1) > 0)
\end{align}

\paragraph{Proof of bad user belief regrets}
The bad user belief case trivially follows the same steps with the additional term:
\begin{align}
    \frac{r_{\hat\pi_t}}{r_t} 
    &= \frac{f(x^*_\text{true}) - f(x_2)}{f(x^*_\text{true}) - f(x_1)}\\
    &\leq \frac{2\beta^{1/2}_t \sigma_{f_{t-1}, \hat{\pi}_{t-1}}(x_2)}{2\beta^{1/2}_t \sigma_{f_{t-1}}(x_1)} + \frac{\lvert \mu_{f_{t-1}}(x_1) -  \mu_{f_{t-1}, \hat{\pi}_{t-1}}(x_2) \rvert}{2\beta^{1/2}_t \sigma_{f_{t-1}}(x_1)} &&\text{(Bad user belief definition)}\\
    &= \Delta \mu_t + R_{\hat\pi_t}
\end{align}
These proofs are for finite decision sets. We can extend this proof to a general decision set by following \citet{srinivas2009gaussian} steps. We omit this procedure, but it essentially boils down to the same procedure and similar results with slight coefficient differences.

\paragraph{Proof of No Harm Guarantee}
We first review the general large $t$ limit properties of $\pi$-augmented GP.
\begin{lemma}
    At the $t \rightarrow \infty$ limit, the posterior GP is asymptotically equal to the original GP.
    \begin{align}
        \lim_{t \rightarrow \infty} \sigma^2_{f_t, \hat{\pi}_t}(x) &= \sigma^2_{f_t}(x),\\
        \lim_{t \rightarrow \infty} \mu_{f_t, \hat{\pi}_t}(x) &= \mu_{f_t}(x),\\
        \lim_{t \rightarrow \infty} \alpha_{f_t, \hat{\pi}_t}(x) &= \alpha_{f_t}(x),\\
        \lim_{t \rightarrow \infty} x_2 &= x_1,
    \end{align}
\end{lemma}
\paragraph{Proof of Lemma 6}
\begin{align}
    \lim_{t \rightarrow \infty} \sigma^2_{f_t, \hat{\pi}_t}(x)
    &= \lim_{t \rightarrow \infty} \frac{\sigma^2_{\hat\pi_t}(x)\sigma^2_{f_t}(x)}{\sigma^2_{\hat\pi_t}(x) + \sigma^2_{f_t}(x)}\\
    &= \lim_{t \rightarrow \infty}
    \frac{(\rho^2(\mathbb{V}_{g_t}[\hat\pi_{g_t}(x)]) + \gamma t^2 \sigma^2_{f_t}(x))\sigma^2_{f_t}(x)}{\rho^2(\mathbb{V}_{g_t}[\hat\pi_{g_t}(x)]) + \gamma t^2 \sigma^2_{f_t}(x) + \sigma^2_{f_t}(x)}\\
    &= \lim_{t \rightarrow \infty}
    \frac{(\rho^2(\mathbb{V}_{g_t}[\hat\pi_{g_t}(x)])/t^2 + \gamma \sigma^2_{f_t}(x))\sigma^2_{f_t}(x)}{\rho^2(\mathbb{V}_{g_t}[\hat\pi_{g_t}(x)])/t^2 + \gamma \sigma^2_{f_t}(x) + \sigma^2_{f_t}(x)/t^2}\\
    &= \frac{\gamma  \sigma^2_{f_t}(x) \sigma^2_{f_t}(x)}{\gamma \sigma^2_{f_t}(x)}\\
    &= \sigma^2_{f_t}(x)
\end{align}
\begin{align}
    \lim_{t \rightarrow \infty} \mu_{f_t, \hat{\pi}_t}(x)
    &= \lim_{t \rightarrow \infty} \frac{\sigma^2_{f_t, \hat{\pi}_t}(x)}{\sigma^2_{\hat\pi_t}(x)} \mu_{\hat\pi_t}(x) + \lim_{t \rightarrow \infty} \frac{\sigma^2_{f_t, \hat{\pi}_t}(x)}{\sigma^2_{f_t}(x)} \mu_{f_t}(x) \\
    &= \lim_{t \rightarrow \infty} \frac{\sigma^2_{f_t}(x)}{\sigma^2_{\hat\pi_t}(x)} \mu_{\hat\pi_t}(x) + \frac{\sigma^2_{f_t}(x)}{\sigma^2_{f_t}(x)} \mu_{f_t}(x) \\
    &= \lim_{t \rightarrow \infty} \frac{\sigma^2_{f_t}(x)}{\rho^2(\mathbb{V}_{g_t}[\hat\pi_{g_t}(x)]) + \gamma t^2 \sigma^2_{f_t}(x)} \mu_{\hat\pi_t}(x) + \mu_{f_t}(x)\\
    &= \mu_{f_t}(x)\\
    \lim_{t \rightarrow \infty} \alpha_{f_t, \hat{\pi}_t}(x)
    &= \lim_{t \rightarrow \infty} \mu_{f_t, \hat{\pi}_t}(x) + \beta_t^{1/2}  \lim_{t \rightarrow \infty} \sigma^2_{f_t, \hat{\pi}_t}(x)\\
    &= \mu_{f_t}(x) + \beta_t^{1/2} \sigma^2_{f_t}(x)\\
    &= \alpha_{f_t}(x)\\
    \lim_{t \rightarrow \infty} x_2
    &= \mathop\mathrm{argmax}_{x \in \mathcal{X}} \lim_{t \rightarrow \infty} \alpha_{f_t, \hat{\pi}_t} (x)\\
    &= \mathop\mathrm{argmax}_{x \in \mathcal{X}} \alpha_{f_t} (x)\\
    &= x_1
\end{align}

\paragraph{Proof of Lemma 2}
By definition and lemma 6, 
\begin{align}
    \lim_{t \rightarrow \infty} r^\pi_t 
    &= \lim_{t \rightarrow \infty} \lvert f(x^*_\text{true}) - f(x_2) \rvert\\
    &= \lvert f(x^*_\text{true}) - f(x_1) \rvert\\
    &= r_t\\
    \lim_{t \rightarrow \infty} \frac{r_{\hat\pi_t}}{r_t} &= 1.
\end{align}

\section{Related work} \label{sup:related}
\paragraph{Eliciting human knowledge for Bayesian optimzation}
There are four ways of incorporating human knowledge, (a) prior over input space \citep{souza2021bayesian, ramachandran2020incorporating, hvarfner2022pi, cisse2023hypbo}, (b) Hyperprior over function space \citep{hutter2011sequential, snoek2014input, wang2021pre}, (c) multi-fidelity information source \citep{song2019general, huang2022bayesian}, (d) unknown constraints \citep{gelbart2014bayesian, hernandez2015predictive, adachi2024adaptive}. They assume fixed and given prior knowledge; we assume implicit and dynamic one. Only $\pi$BO and ours offer theoretical guarantees for misspecified $\pi$. $\pi$ is a soft constraint, distinct from the hard one in category (d). They overly limit the search space if incorrect.

\paragraph{Human preference as the objective}
Preferential BO (PBO) is to find the sample that maximize the probability to be Condorcet winner via interaction with users \citep{gonzalez2017preferential, astudillo2023qeubo, takeno2023towards}, then extended to preference exploration for multi-objective BO \citep{lin2022preference, astudillo2020multi, taylor2021bayesian, giovanelli2023interactive}, non-duel-based PBO \citep{brochu2007active, koyama2020sequential, mikkola2020projective, benavoli2023bayesian}. 
While they treat preferences as the objective, ours regards it as an additional information source of the original optimization task.

\paragraph{Human-AI teaming BO}
An interactive human-BO collaboration setting has recently been proposed to accommodate time-varying human knowledge. There are two categories: (a) Human rectifies BO suggestions; \citet{colella2020human, av2022human}, (b) BO supports human manual search via exploratory adjustment \citet{gupta2023bo, khoshvishkaie2023cooperative}. While (a) has no guarantee on worst-case, (b) has a worse convergence rate than the standard BO \citep{khoshvishkaie2023cooperative}. We proposed the first approach with a no-harm guarantee in an interactive setting without losing human initiative.

\paragraph{Explainability in BO}
There are two prior works on explainable BO. \citet{li2020explainability} integrate explainability as hard constraints to the optimization that restrict the search space to where humans can interpret. \citet{liu2023sparse} introduced sparsity to find simple and interpretable configurations via L0 regularization. Both methods do not consider human interaction, but these are orthogonal to ours and may be beneficial to combine.

\section{Modelling Human Preference via Gaussian Process} \label{sup:pref}
\paragraph{Preferential Gaussian process modelling}
While many algorithms, rooted in either probabilistic methods~\citep{bradley1952rank, luce1959possible} or spectral approaches~\citep{cucuringu2016sync,chau2022spectral}, can model human preferences, we opted for Gaussian Processes (GP). This choice enables us to consider epistemic uncertainty more effectively during modeling. However, the classical preferential GP~(PGP)\citep{chu2005preference} has several limitations; it is computationally expensive and unable to learn preferences that might be inconsistent and with heteroscedastic noise. We combined two existing simple approaches instead of PGP; Dirichlet-based GP (DGP) \citep{milios2018dirichlet}, and skew-symmetric data-augmentation \citep{chau2022learning}. DGP translates classification problem as regression one via transforming classification labels to the coefficients of a degenerate Dirichlet distribution. This offers a fast and heteroscedastic GP classifier that has essentially the same accuracy and uncertainty quantification as the original GP classifier. Skew-symmetric data augmentation is to add the symmetric data of original data $Y_{j, i} = 1 - Y_{i, j}$ for the duel $(x_i, x_j)$. Additionally, we used Bayesian quadrature for fast approximation of integral in Eq.~\labelcref{eq:pi_int}. However, our setting is not limited to this GP and Bayesian quadrature approximation.

\paragraph{Bayesian quadrature modelling}
\begin{figure*}
  \centering
  \includegraphics[width=\hsize]{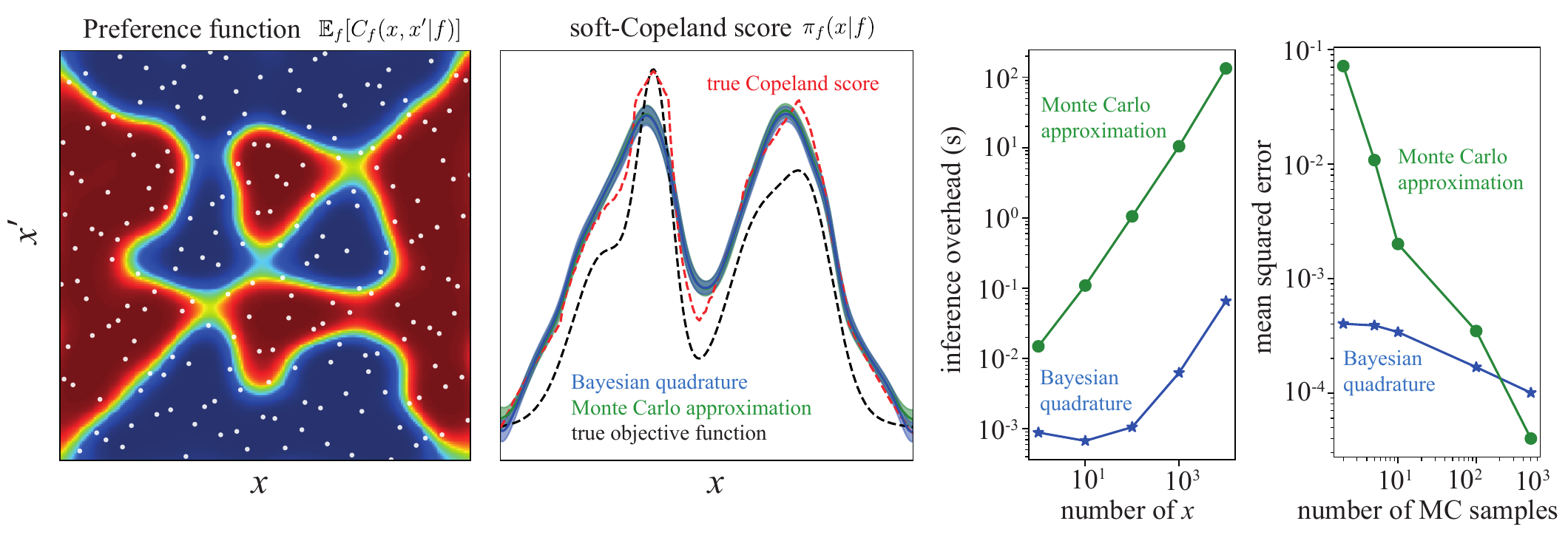}
  \caption{Comparison of Monte Carlo (MC) approximation and Bayesian quadrature approximation on marginalisation for soft-Copeland score. Overhead refers to the wall-clock time to compute the soft-Copeland score at $x$ with the 100 MC samples. The mean squared error is the error between the estimated soft-Copeland score and the ground truth computed with massive MC samples.}
  \label{fig:bq}
  \vspace{-1em}
\end{figure*}
As the integration Eq.~(\ref{eq:pi_int}) is intractable, we have to approximate it. The original preferential BO \citep{gonzalez2017preferential} adopts Monte Carlo (MC) approximation, but it is slow as the authors admit there should be a better way. We adopted Bayesian quadrature (BQ) \cite{o1991bayes}, particularly BASQ \citep{adachi2022fast, adachi2023bayesian} for fast approximation. Figure \ref{fig:bq} compares the overhead and accuracy of MC integration and BQ. BQ yields faster computation than MC samples at inference (BQ: $\mathcal{O}((n_\text{MC} + n_\text{duel})^2)$ vs. MC: $\mathcal{O}(n_\text{MC} n_\text{func} n_\text{duel}^2)$), which repeats the computation in the acquisition function optimization loop. Of course, training the BQ model takes additional cost ($\mathcal{O}(n_\text{mBCG} n_\text{duel}^2)$ for mBCG algorithm \citep{gardner2018gpytorch}), but this can be understood as "pretraining" to avoid the quartic cost of MC integration at each inference point. As RBF kernel assumption of BQ is misspecified against true Bernoulli distribution, the convergence over sample size is limiting; still, BQ works well as it is robust against misspecification \citep{kanagawa2016convergence}. In larger MC sample sizes, MC integration outperforms BQ, but it produces non-negligible overhead. It should be noted that BQ can derive the closed form of the denominator of Eq.~(\ref{eq:pi_int}), whereas MC approximation requires another higher-level MC approximation of quintic cost, which is prohibitively slow. So, we adopt BQ in this paper thanks to its good balance of computational accuracy and cost; however, our setting is not limited to BQ.

\section{Bayesian Quadrature for Fast Soft-Copeland Score Approximation} \label{sup:BQ}
We have GP classifier $g \sim \mathcal{GP}(\mu_g, \kappa_g)$, and Monte Carlo integration via transforming sampled function with the link function can estimate the expectation of binary probability as Bernoulli distribution:
\begin{align}
    \mathbb{E}_g[g(x, x^\prime | g, \textbf{D}_\text{pref})] = \int \frac{\exp(g_i)}{\sum \exp(g_i)} \mathbb{P}(g_i | x, x^\prime, \textbf{D}_\text{pref}) \text{d} g
\end{align}
As this is intractable, the forthcoming Condorset winner $\hat\pi_g(x)$ marginalisation is also intractable.

Therefore, we apply Bayesian quadrature (BQ). Bayesian quadrature is the model-based approximation technique for intractable integration; typically, GP is applied for the surrogate model for the integrand. We apply the simple GP with RBF kernel to the integrand. Namely, we place the surrogate model GP with the pair of dataset:
\begin{align}
    \textbf{x}_\text{bq} &:= \textbf{x}_\text{pref} = (\textbf{x}_1, \textbf{x}_2), \\
    \textbf{y}_\text{bq} &:= \mathbb{E}_g[g(\textbf{x}_\text{pref} \mid \textbf{D}_\text{pref})], \\
    \textbf{D}_\text{bq} &:= \left(\textbf{x}_\text{bq},  \textbf{y}_\text{bq} \right),
\end{align}

Then, all integration in \eqref{eq:pi_int} becomes analytical:
\begin{align}
    f_\text{bq} &\sim \mathcal{GP}(\mu_\text{bq}, \kappa_\text{bq} \mid \textbf{D}_\text{bq}), \\
    \mu_\text{bq}(X) &= k(X, \textbf{x}_\text{bq}) \left[ k(\textbf{x}_\text{bq}, \textbf{x}_\text{bq}) + \lambda \textbf{I} \right]^{-1} \textbf{y}_\text{bq}, \\
    &= k(X, \textbf{x}_\text{bq})^\top \boldsymbol\omega,\\
    &= v^\prime \mathcal{N} (X; \textbf{x}_\text{bq}, \textbf{W})^\top \boldsymbol\omega,\\
    \kappa_\text{bq}(X, X^\prime) &= k(X, X^\prime) - k(X, \textbf{x}_\text{bq}) \left[ k(\textbf{x}_\text{bq}, \textbf{x}_\text{bq}) + \lambda \textbf{I} \right]^{-1} k(\textbf{x}_\text{bq}, X),\\
    &= v^\prime \mathcal{N} (X; x^\prime, \textbf{W}) - v^{\prime \, 2} \mathcal{N} (X; \textbf{x}_\text{bq}, \textbf{W}) \boldsymbol\Omega \mathcal{N} (X; \textbf{x}_\text{bq}, \textbf{W})^T, \\
    &= v^\prime \mathcal{N} (X; x^\prime, \textbf{W}) - v^{\prime \, 2} \sum_{i,j}^{N} \Omega_{i,j} \mathcal{N} (X; x_{bq, i}, \textbf{W}) \mathcal{N} (X; x_{bq, j}, \textbf{W}), \\
    &= v^\prime \mathcal{N} (X; X^\prime, \textbf{W}) - v^{\prime \, 2} \mathcal{N} (x_{bq, i}; x_{bq, j}, 2 \textbf{W}) \sum_{i,j}^{N} \Omega_{i,j} \mathcal{N} \left(X; \frac{x_{bq, i} + x_{bq, j}}{2}, \frac{\textbf{W}}{2} \right),
\end{align}
where
\begin{align}
    X &:= (x, x^\prime),\\
    k(X, X^\prime) &:= v^\prime \mathcal{N} (X; X^\prime, \textbf{W}), \\
    v^\prime &= v\sqrt{\lvert 2 \pi \textbf{W} \rvert}, \\
    \textbf{W} &:= \ell^2 \textbf{I}, \\
    \boldsymbol\omega &:= \left[ k(\textbf{x}_\text{bq}, \textbf{x}_\text{bq}) + \lambda \textbf{I} \right]^{-1} \textbf{y}_\text{bq}, \\
    \boldsymbol\Omega &:= \left[ k(\textbf{x}_\text{bq}, \textbf{x}_\text{bq}) + \lambda \textbf{I} \right]^{-1},
\end{align}
$v$ and $\ell$ are the output scale and length scale of the RBF kernel, and $\lambda$ is the Gaussian likelihood variance. Posterior predictive mean and variance are just a mixture of Gaussians; thus integration is tractable.

Then, we consider the soft-Copeland score, which is the marginalisation of $x^\prime$ from $x = (x, x^\prime)$. Marginalisation of Gaussian is just extracting the corresponding elements. We have
\begin{align}
    X = \begin{bmatrix} x\\ x^\prime \end{bmatrix} \qquad
    \textbf{x}_\text{bq} = \begin{bmatrix} x_\text{bq}\\ x_\text{bq}^\prime \end{bmatrix} \qquad
    \textbf{W} = \begin{bmatrix} \textbf{W}^\prime & \textbf{0} \\  \textbf{0} & \textbf{W}^{\prime\prime} \end{bmatrix} \qquad
\end{align}
Consequently, the soft-Copeland score is reduced to be:
\begin{align}
    \hat\pi_g(x) &:= \int_\mathcal{X} g(x, x^\prime) \text{d} x^\prime,\\
    &= \text{V}_\mathcal{X}^{-1} \int g(x, x^\prime) \text{d} x^\prime,
\end{align}
where $\text{V}_\mathcal{X} = \int \mathbb{E}_g [\hat\pi_g(x)] \text{d} x$ is the normalizing constant to make $\hat\pi_g(x)$ become a probability density function.
\begin{align}
    \mathbb{E}_g \left[ \hat\pi_g(x) \right] &= \text{V}_\mathcal{X}^{-1} \int \mathbb{E}_g \left[g(x, x^\prime) \right] \text{d} x^\prime,\\
    &= \text{V}_\mathcal{X}^{-1} \int \mu_\text{bq}(x, x^\prime) \text{d} x^\prime, \\
    &= v^\prime \text{V}_\mathcal{X}^{-1} \int \mathcal{N} \left(
    X; \textbf{x}_\text{bq}, \textbf{W}
    \right)^\top \text{d} x^\prime \boldsymbol\omega, \\
    &= v^\prime \text{V}_\mathcal{X}^{-1} \int \mathcal{N} \left(
    \begin{bmatrix} x\\ x^\prime \end{bmatrix};
    \begin{bmatrix} x_\text{bq}\\ x_\text{bq}^\prime \end{bmatrix}, 
    \begin{bmatrix} \textbf{W}^\prime & \textbf{0} \\  \textbf{0} & \textbf{W}^{\prime\prime} \end{bmatrix}
    \right)^\top \text{d} x^\prime \boldsymbol\omega, \\
    &= v^\prime \text{V}_\mathcal{X}^{-1} \mathcal{N} (x; x_\text{bq}, \textbf{W}^\prime)^\top \boldsymbol\omega,
\end{align}
Similarly, we place another GP on the variance,
$\mathbb{V}_g \left[ g(x, x^\prime) \right]$, then the procedure is the same:
\begin{align}
    \mathbb{V}_g \left[ \hat\pi_g(x) \right] &= \text{V}_\mathcal{X}^{-1} \int \mathbb{E}_g \left[
    g(x, x^\prime) 
    (1 - g(x, x^\prime ))
    \right] \text{d} x^\prime,\\
    &\approx \text{V}_\mathcal{X}^{-1} \int \mu^\prime_\text{bq}(x, x^\prime) \text{d} x^\prime, \\
    &= v^\prime \text{V}_\mathcal{X}^{-1} \mathcal{N} (x; x^{\prime \prime}_\text{bq}, \textbf{W}^{\prime \prime \prime})^\top \boldsymbol\omega,
\end{align}
Then, the soft-Copeland score can be approximated by moment-matching the original Bernoulli distribution with the Gaussian distribution.
\begin{align}
    \pi_g(x) \approx \mathcal{N}(\hat\pi_g(x); \mathbb{E}_g \left[ \hat\pi_g(x) \right], \mathbb{V}_g \left[ \hat\pi_g(x) \right]).
\end{align}
This is a coarse approximation of Condorcet winner probability $\mathbb{P}(y = 1 | x)$ and is probabilistically wrong (Gaussian is not bounded in [0, 1]). Still, recall our original motivation is to model the probability of global optimal location, which is not bounded in [0, 1]. In this sense, precisely computing Bernoulli distribution is not required. We further use this soft-Copeland score function to model the prior distribution on the continuous value $y$, which is not the Bernoulli distribution and rather assumes Gaussian distribution. Thus we adopt Gaussian moment-matching approximation.

However, this soft-Copeland score is not normalised over the domain, so we need to take further integral over $x$ domain. Bayesian quadrature makes this integral analytical:
\begin{align}
    \text{V}_\mathcal{X} &= \int_\mathcal{X} \mathbb{E}_g \left[ \hat\pi_g(x) \right] \text{d} x\\
    &= v^\prime \text{V}_\mathcal{X}^{-1} \int \mathcal{N} (x; x_\text{bq}, \textbf{W}^\prime)^\top  \text{d} x \boldsymbol\omega,\\
    &= v^\prime \text{V}_\mathcal{X}^{-1} \textbf{1}^\top \boldsymbol\omega,\\
    &= \sqrt{v^\prime \textbf{1}^\top \boldsymbol\omega},
\end{align}
\begin{align}
    \pi(x) &:= \frac{\mathbb{E}_g[\pi_g(x]}{\int_\mathcal{X} \mathbb{E}_g[\hat\pi_g(x)] \text{d} x}, \\
    &= \frac{1}{\textbf{1}^\top \boldsymbol\omega} \mathcal{N} (x; x_\text{bq}, \textbf{W}^\prime)^\top \boldsymbol\omega
\end{align}

Now we have the closed-form soft-Copeland score approximation. It should be noted that the original $g(x, x^\prime)$ distribution is the Bernoulli distribution, whereas this BQ-GP is the Gaussian distribution, which is a crude approximation. So only predictive mean is reliable. To estimate the variance of $\mathbb{V}_g[\hat\pi_g(x)]$ at the same time, we need to have another GP for variance estimation.
Even the predictive mean of BQ-GP can be a crude approximation. One simple solution to boost the accuracy is to increase the number of data $\textbf{D}_\text{bq}$, as this can be augmented cheaply by $f_\text{pref}$. However, increasing the number can lead to large computational overhead as training GP costs cubic complexity $\mathcal{O}(n^3)$. We wish to minimise the number of augmented data. So we adopt BASQ \cite{adachi2022fast} for selecting the next query point. This allows provably small predictive uncertainty (see Theorem 1 in \cite{adachi2022fast} and also \cite{hayakawa2022positively} ).

\section{Dueling Acquisition Function} \label{sup:fbpibo}
\begin{figure*}
  \centering
  \includegraphics[width=\hsize]{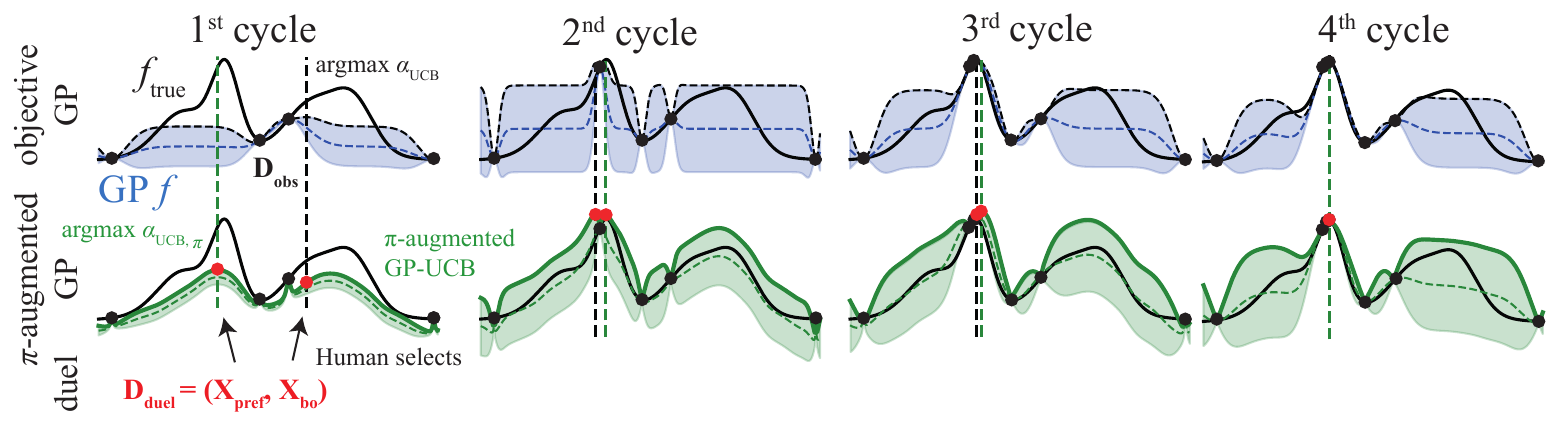}
  \caption{Dueling candidate generation algorithm. While the optimistic sample is selected by maximizing the $\pi$-augmented GP, the pessimistic sample is selected by the original objective GP. The distance between pairwise samples asymptotically decreases over iterations. (We set $\gamma = 0.1$).}
  \label{fig:demo}
\end{figure*}

Figure~\ref{fig:demo} visualizes the dueling acquisition function. In the first cycle, the distance between preference-based and standard UCB-based recommendations is large. But it gradually decreases over iterations, and it is almost the same in the last (fourth) iteration. Furthermore, this figure also shows that the human preference successfully avoids climbing up the wrong left peak by placing the belief over the left peak, which accelerates early-stage exploration.

\section{Explaining Bayesian Optimization}\label{sup:explain}
We have three explanation features explained in Figure~\ref{fig:explanation}; Spatial relation, feature importance, and selection accuracy feedback. 
For the spatial relation, we select the two primary dimensions using Shapley values. As Shapley values are conditioned on $x$, we need to select $x$. For simplicity, we take average of Shapley values at the pairwise candidates, $x_1$ and $x_2$, then take two dimensions with top 2 mean Shapley values.
For the drawing range, we first compute the rectangle which bounds the following three points; $x_1$, $x_2$, and the current best observed point $x_\text{current} := \arg \max {\textbf{D}_{obj}}_t$. Then, we expand this bounded rectangle 2 times as large as the original for visibility. We set this expanded rectangle as the visualisation range. This procedure is shared with the preference model visualisation.
For feature importance, we simply visualise the Shapley values at $x_1$ and $x_2$ as a bar plot.

For selection accuracy feedback, we first update the GP surrogate function with the queried point $x_t \in (x_1, x_2)$ and the queried value $y_t = f(x_t)$. For the sake of argument, we assume $x_t = x_1$. Then, we compute the probability of correct selection, given by:
\begin{align}
\mathbb{P}(f(x_1) \geq f(x_2)) &\sim \mathcal{N}(\mathbb{E}_f[\ell(x_1, x_2)], \mathbb{V}_f[\ell(x_1, x_2)]),\\
\ell(x_1, x_2) &:= \Phi \left(\frac{f(x_1) - f(x_2)}{\sqrt{\lambda}} \right),
\end{align}
where $\Phi$ is the cumulative density function of standard normal distribution $\mathcal{N}(0,1)$. We compute the expectation and variance over $f$ space by Monte Carlo integration.

\section{Experimental details} \label{sup:exp}
We have tested CoExBO for 5 synthetic functions against 6 baselines.
We use a constant-mean GP with an RBF kernel. In each iteration of the active learning loop, the outputs are standardized to have zero mean and unit variance. We optimize the hyperparameter by maximizing the marginal likelihood (type-II maximum likelihood estimation) using L-BFGS-B optimizer \citep{liu1989limited} implemented with BoTorch \citep{balandat2020botorch}. We also maximized the acquisition function using the same optimizer. The initial data sets consist of ten data points drawn by Sobol sequence \citep{sobol1967distribution} and corresponding noisy observations. We generate 10 samples for objective dataset and 100 samples for preferential dataset construction.
We adopt the log regret as the evaluation metric using the test dataset. The models are implemented in GPyTorch \citep{gardner2018gpytorch}. All experiments are repeated ten times with different initial data sets via different random seeds (the seeds are shared with baseline methods). 

\subsection{Synthetic functions with Synthetic Human Selection} \label{sup:synthetic}
\subsubsection{Synthetic Functions}
\paragraph{Ackely}
Ackley funciton is defined as:
\begin{align}
f(x) := - a \exp \left[ -b \sqrt{\frac{1}{d} \sum_{i=1}^d x_i^2} \right] - \exp \left[
\frac{1}{d} \sum_{i=1}^d \cos (c x_i) \right] + a + \exp(1)
\end{align}
where $a = 20, c = 2\pi, d = 4$. We take the negative Ackley function as the objective of BO to make this optimisation problem maximisation. This is a 4-dimensional function bounded by $x \in [-1, 1]^d$. The global optimum is $x^*_\text{true} = [0,0,0,0]$ and $f(x^*_\text{true}) = 0$.

\paragraph{Hölder Table}
Hölder Table funciton is defined as:
\begin{align}
f(x) := \Bigg\lvert \sin(x_1) \cos(x_2) \exp\left( \Bigg\lvert 1 - \frac{\sqrt{x_1^2 + x_2^2}}{\pi} \Bigg\rvert \right) \Bigg\rvert
\end{align}
where $x_i$ is the $i$-th dimensional input. This is a 2-dimensional function bounded by $x \in [0, 10]^d$. The global optimum is $x^*_\text{true} = [8.05502, 9.66459]$ and $f(x^*_\text{true}) = 19.2085$.

\paragraph{Styblinski-Tang}
Styblinski-Tang funciton is defined as:
\begin{align}
f(x) := \frac{1}{2} \sum_{i=1}^d (x_i^4 - 16 x_i^2 + 5x_i)
\end{align}
where $x_i$ is the $i$-th dimensional input. This is a 3-dimensional function bounded by $x \in [-5, 5]^d$. The global optimum is $x^*_\text{true} = [-2.903534]^d$ and $f(x^*_\text{true}) = 39.166166 d$.

\paragraph{Michalewicz}
Michalewicz funciton is defined as:
\begin{align}
f(x) := \sum_{i=1}^d \sin(x_i) \sin^{2m} \left(\frac{i x_i^2}{\pi} \right)
\end{align}
where $x_i$ is the $i$-th dimensional input and $m =10$. This is a 5-dimensional function bounded by $x \in [0, \pi]^d$. The global optimum is  $f(x^*_\text{true}) = 4.687658$.

\paragraph{Rosenbrock}
Rosenbrock funciton is defined as:
\begin{align}
f(x) :=  \sum_{i=1}^{d-1} \left[ 100 (x_{i+1} - x_i^2)^2 + (x_i - 1)^2 \right]
\end{align}
where $x_i$ is the $i$-th dimensional input. This is a 3-dimensional function bounded by $x \in [-5, 10]^d$. The global optimum is $x^*_\text{true} = [1]^d$ and $f(x^*_\text{true}) = 0$.

\subsubsection{Robustness evaluation}
In the adversarial selection, both CoExBO and batchUCB superficially surpass the original UCB. We first point out that this is within the standard error, but there may be possible reasons why these two are better than others even in adversarial settings. For CoExBO, this may come from randomised effect. Recent work shows randomizing $\beta$ parameter of UCB yields faster convergence than original \citep{berk2020randomised, takeno2023randomized}. Our CoExBO can be understood as randomising $\beta$, which may effect positively. Still, we can confirm the trend that confident and correct human feedback can accelerate convergence. For batchUCB, this may come from a nonmyopic effect. \citet{gonzalez2016glasses} pointed out the similarity between hallucination and one-step lookahead BO, which empirically yields better convergence than the original UCB.

\subsection{Real-world tasks}
\begin{figure}
  \centering
  \includegraphics[width=0.6\hsize]{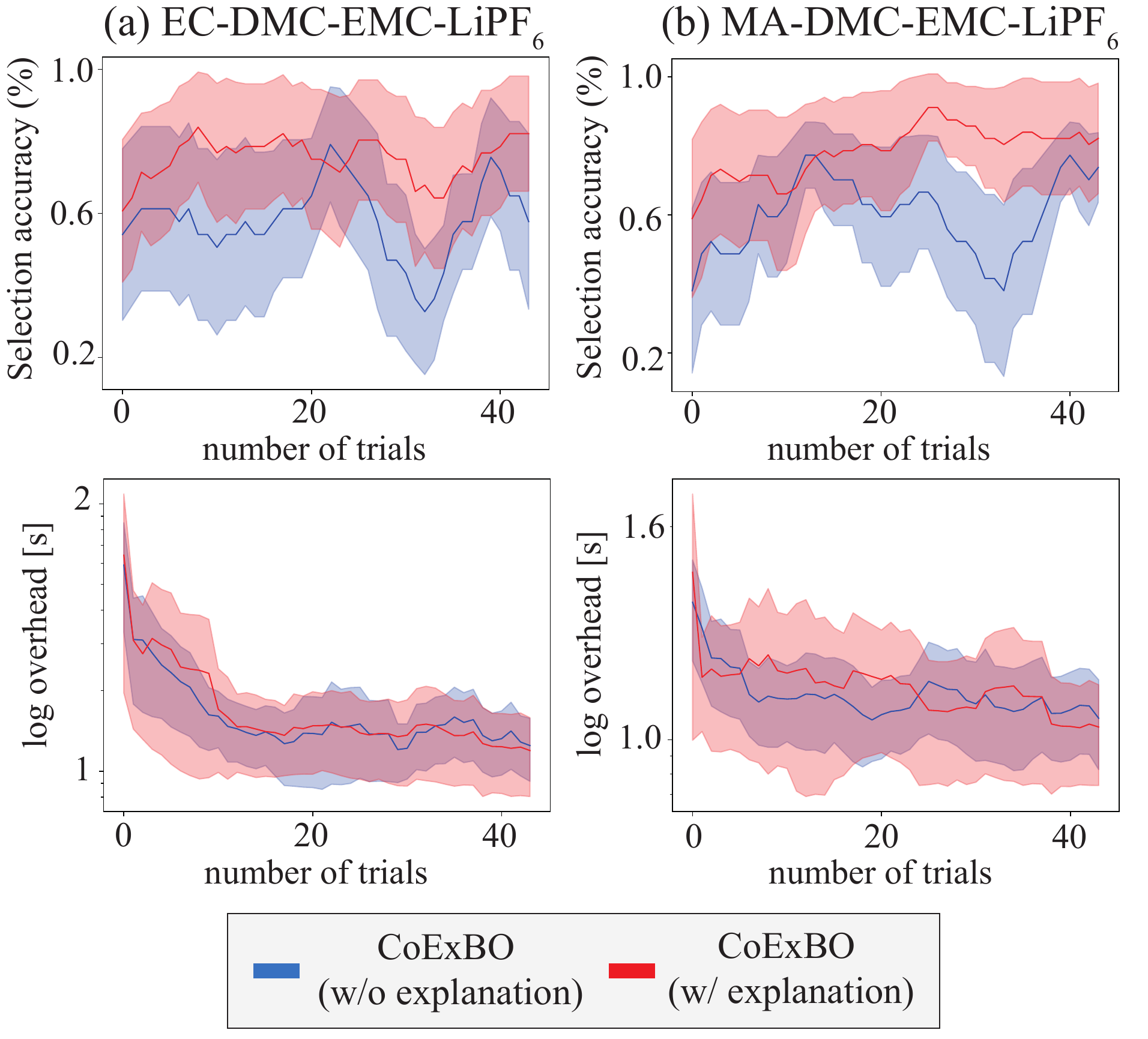}
  \caption{Selection accuracy and overhead analysis over iterations for the tasks (a) EC-DMC-EMC-LiPF$_6$ and (b) MA-DMC-EMC-LiPF$_6$. The solid lines and shaded areas refer to the mean and 1 standard error of the results of four participants. To smooth out the noisy results, we take the moving average with the window size of 3 trials for visualizing the trend.}
  \label{fig:complexity}
\end{figure}

\subsubsection{Designing battery}\label{sup:battery}
The problem involves finding the best electrolyte material combination to maximize ionic conductivity. Ionic conductivity is dependent on both lithium salt molarity and the cosolvent composition. They show the complex non-linear relationship due to the solvation effect and cannot predict even with the state-of-the-art quantum chemistry simulator. We create the true functions by fitting the experimental data of EC-DMC-EMC-LiPF$_6$ \citep{dave2022autonomous} and MA-DMC-EMC-LiPF$_6$ \citep{logan2018study} systems using the Casteel-Amis equation \citep{casteel1972specific}. Note that the Casteel-Amis equation is just for the interpolation of experimental data to be continuous, and is not capable of predicting different cosolvent. Both tasks are the three-dimensional continuous input function.
The input features are (1) the lithium salt (LiPF$_6$) molarity, (2) DMC/EMC cosolvent ratio, and (3) (EC or MA)/carbonates cosolvent ratio, respectively. The inputs are bounded with $x_1 \in [0, 2]$, $x_2 \in [0, 1]$, and $x_3 \in [0, 1]$. The noisy output is generated by adding i.i.d. zero-mean Gaussian noise with the $3^2$ variance to the noiseless $f(x)$. 

\subsubsection{Selection Accuracy and Complexity Analysis}
We further analyzed the real-world task results based on selection accuracy and overhead over iterations. Figure~\ref{fig:complexity} illustrates the results. For selection accuracy, while results with explanation remain stable over iterations, the ones without explanation fluctuate largely, particularly in the later rounds. Over iterations, the pairwise candidates become closer due to the no-harm guarantee. Hence, the later iterations are more difficult to select the correct one. The explanation feature can help users distinguish the slight differences by the quantitative Shapley values, leading to accurate selection even for the later iterations.
The bottom row of  Figure~\ref{fig:complexity} shows the overhead of the candidate selection process, including pairwise candidate generation, explanation feature, and human selection time. We can observe the general decrease trend over iterations regardless of the explanation feature, and the difference in overhead between with and without explanation features is negligible. This is because the most time-consuming part is the human selection process. In the early stage, human users are also uncertain and need more time to decide which to select. Over time, it becomes more confident and smoother to select, resulting in quicker selection. This suggests the algorithmic overhead is negligible when compared to the human selection process. 

\end{document}

%% file: math_commands.tex
\usepackage{amsmath,amsfonts,bm}

\def\eqref#1{equation~\ref{#1}}
\def\1{\bm{1}}

\DeclareMathAlphabet{\mathsfit}{\encodingdefault}{\sfdefault}{m}{sl}
\SetMathAlphabet{\mathsfit}{bold}{\encodingdefault}{\sfdefault}{bx}{n}

\DeclareMathOperator*{\argmax}{arg\,max}

%% file: math_alan.tex
\def\bfB{{\bf B}}

\def\bff{{\bf f}}

\def\RR{{\mathbb R}}    %r�els
\def\PP{{\mathbb P}}     %espace projectif / probabilité
\def\EE{{\mathbb E}}    % espérance
\def\VV{{\mathbb V}}
\def\11{{\mathbf 1}}    % indicatrice

  \def\cG{{\mathcal G}}      \def\cN{{\mathcal N}}          \def\cP{{\mathcal P}}           \def\cX{{\mathcal X}}   

\def\bfx{{\bf x}} \def\bfy{{\bf y}}  
 \def\bfK{{\bf K}}

\def\bfX{{\bf X}}